\documentclass[11pt,a4paper]{article}
\usepackage{times,latexsym}
\usepackage{url}
\usepackage[T1]{fontenc}

\usepackage{algorithm} % For floating container
\usepackage{algpseudocode} % For the actual type setting https://tex.stackexchange.com/questions/229355/algorithm-algorithmic-algorithmicx-algorithm2e-algpseudocode-confused
\usepackage{graphicx}
\usepackage{bm} %\bm
\usepackage{amssymb} %\mathbb
\usepackage{amsmath} %\text
\usepackage{tikz} %\mathbb
\usetikzlibrary{shapes,shapes.multipart,fit,external,backgrounds,patterns}
\usepackage[frozencache,cachedir=.]{minted}
\newcommand{\argmax}{\mathop{\mbox{argmax}}}

\usepackage[acceptedWithA]{tacl2021v1}
\usepackage{xspace,mfirstuc,tabulary}

\newif\iftaclinstructions
\taclinstructionsfalse % AUTHORS: do NOT set this to true
\iftaclinstructions
\renewcommand{\confidential}{}
\renewcommand{\anonsubtext}{(No author info supplied here, for consistency with
TACL-submission anonymization requirements)}
\newcommand{\instr}
\fi

\iftaclpubformat % this "if" is set by the choice of options

\else

\fi

\title{Segmentation-Free Streaming Machine Translation}

\author{
   Javier Iranzo-Sánchez$^\diamond$ 
   \and
   Jorge Iranzo-Sánchez$^\diamond$
   \and
   Adrià Giménez$^\dagger$
   \\
   {\bf Jorge Civera$^\diamond$}
   \and
   {\bf Alfons Juan$^\diamond$}
   \\
   \ \\
   $^\diamond$Machine Learning and Language Processing, VRAIN, Universitat Politècnica de València
   \\
   \texttt{\{jairsan,jorirsan,jorcisai,ajuanci\}@upv.es}
   \ \\
   \\
   $^\dagger$Departament d'Informàtica, Escola Tècnica Superior d'Enginyeria, Universitat de València
   \\
   \texttt{adria.gimenez@uv.es}
 }

\date{}

\begin{document}
\maketitle
\begin{abstract}
Streaming Machine Translation (MT) is the task of translating an
unbounded input text stream in real-time. The traditional cascade
approach, which combines an Automatic Speech Recognition (ASR) and an
MT system, relies on an intermediate segmentation step which splits
the transcription stream into sentence-like units. However, the
incorporation of a hard segmentation constrains the MT system and is a
source of errors. This paper proposes a Segmentation-Free framework
that enables the model to translate an unsegmented source stream by
delaying the segmentation decision until after the translation has
been generated. Extensive experiments show how the proposed
Segmentation-Free framework has better quality-latency trade-off than
competing approaches that use an independent segmentation model.
	\footnote{Software, data and models are available at \url{https://github.com/jairsan/Segmentation-Free_Streaming_Machine_Translation}}
\end{abstract}

\section{Introduction}
Streaming Machine Translation (STR-MT) is a specific task of Machine
Translation (MT) that consists in translating an unbounded input text
stream in real-time. STR-MT systems are typically used in a cascaded
setting following a streaming Automatic Speech Recognition (ASR)
system. This task has many applications for scenarios such as live
broadcasting, parliamentary debates, live lectures, etc.  where the
input speech to be translated is potentially several hours long, and
the system must provide accurate and real-time translations over the
live session.

However, conventional MT systems are not well prepared to work in the
conditions described above. Training samples for conventional MT
systems are sentence-aligned pairs, so there is a length mismatch
between the training (a few hundred tokens at most) and inference
conditions (thousands of tokens for live sessions).  Conventionally,
some sort of segmentation model is used in order to split the incoming
text stream into sentence-like units, so that they can be translated
by the MT system. Each sentence-like unit, or segment, is typically
translated in isolation, although techniques from document MT can be
used to provide additional context to a conventional MT model beyond
the current
sentence~\cite{tiedemann-scherrer-2017-neural,agrawal-etal-2018-contextual,scherrer-etal-2019-analysing,ma-etal-2020-simple,Zheng2020b,Li2020b,zhang-etal-2021-beyond}. These
techniques can be adapted to the streaming case using the concept of
streaming history~\cite{iranzo-sanchez-etal-2022-simultaneous}, which
keeps a limited context of the previously seen source segments and
their corresponding translations generated by the MT model. We
collectively refer to all approaches that use an independent upstream
segmenter model as the {\it Segmented setting}.

The main downside of the Segmented setting is that the translation
quality is very dependant on the quality of the segmenter, and forcing
a hard decision without involving the MT system conditions the
resulting translation quality. This is particularly relevant in
scenarios where we have a downstream system. On this line of work, our
final goal is to build a streaming speech-to-speech translation system
using an additional downstream streaming TTS model. The decisions of
the TTS model cannot be changed once the output has been sent to the
user, and therefore it is not possible to change the output of the
segmenter/MT systems.

This paper proposes a Segmentation-Free (SegFree) approach that does
not rely on an upstream segmenter.  Instead, the MT model receives an
unsegmented stream of source text and generates a continuous sequence
of translated words.  The SegFree model jointly generates the
translation and its target segmentation by inserting a special token
(``[SEP]'') into the translation stream.  Whereas in the Segmented
setting, the segmentation decision is incorporated into the source
side independently from the MT model, in the SegFree setting this
decision is taken by the MT model considering both, the source and
target streams. That is, the segmentation decision has been moved from
an upstream segmenter into the target translation stream. Delaying
this segmentation decision allows the SegFree model to take into
account additional target-side information, which is the crucial
component that enables the SegFree model to significantly outperform
its Segmented counterpart.

The rest of this article is organised as follows. First, the related
work is described in Section~\ref{sec:related}. Then,
Section~\ref{sec:framework} defines the statistical framework of our
proposed Segmentation-Free model. Next, Section~\ref{sec:settings}
introduces the datasets involved, how they were processed and the
instantiation of the models. Lastly, Section~\ref{sec:results} reports
the results achieved and conclusions are drawn in
Section~\ref{sec:conclusions}.

\section{Related work}
\label{sec:related}

The problem of automatic segmentation has limited the MT community for
many years, and many solutions have been proposed, ranging from simple
length-based heuristics~\cite{Cettolo2006}, using language model
probabilities~\cite{Stolcke1996}, to introducing segmentation with a
monolingual MT
system~\cite{cho-etal-2012-segmentationb,cho-etal-2015-punctuation,Cho2017}.
\citet{li2021} proposes a data augmentation technique that introduces
segmentation errors during training in order to make the model more
robust to this type of errors.

MT of unsegmented inputs has received relatively little
attention. \citet{KolssVW08}~propose a stream decoding algorithm for
phrase-based statistical MT, that is able to translate unsegmented
input by using a continuous translation lattice that is updated during
the translation process. New input words extend the lattice, and the
output is committed whenever a predefined latency is exceeded. In the
case of neural MT systems, \citet{schneider-waibel-2020-towards} use a
Transformer-XL~\cite{dai-etal-2019-transformer} encoder for longer
context, combined with a monotonic encoder-decoder attention head. The
model translates unsegmented input text using a rolling window over
the source stream with a fixed offset. Their training procedure is a
multi-stage method that involves multiple training phases. When
translating in a streaming setup, the latency of the model is quite
significant when compared with the
speaker~\cite{iranzo-sanchez-etal-2022-simultaneous}.  More recently,
\citet{sukanta2022} propose a method for the translation of
unsegmented input by using a small window over the source stream that
is re-translated each time a new input token is received. The
overlapping translations of each window are then merged together to
form the output stream. The downside of this approach is that it
introduces flickering into the output, so it is difficult to apply to
setups such as speech-to-speech translation that require
non-flickering output.

%In contrast to this, the more general task of improving the
%translation policy of simultaneous systems has received comparatively
%more attention. 

With regards to the translation models themselves, a distinction can
be made between those that use fixed policies, policies which are
pre-defined and do not depend on the current content being translated,
and adaptive policies, which are those that adapt their decision based
on the current translation status. Wait-$k$~\cite{ma-etal-2019-stacl}
is the most popular fixed translation policy. This policy first waits
for $k$ source words to arrive, and then alternates between writing a
new target word and waiting for a new source word. Adaptive policies
are much more varied, although they can be classified along some
general trends. One popular approach is to try to first detect
meaningful units or chunks that must be translated together, and only
generate a translation once an entire chunk has been
received~\cite{wilken-etal-2020-neural,zhang-etal-2020-learning-adaptive,
  kano-etal-2021-simultaneous, bahar-etal-2021-without,
  zhang-etal-2022-learning}.  In contrast, in some approaches the
policy is derived from the model itself, either from the output
probabilities~\cite{ChoE16,zheng-etal-2020-simultaneous,liu-etal-2021-cross}
or from some internal state of the
model~\cite{arivazhagan-etal-2019-monotonic, MaPCPG20}.

\section{Segmentation-Free Statistical Framework}
\label{sec:framework}

Under the proposed SegFree framework, the translation system receives
an unsegmented, continuous stream of source words, and produces a
translation stream in a real-time fashion. Unlike in the Segmented
setting, the system is not constrained by the pre-existing
segmentation, so it decides how to delimit the segments of the output
stream by taking into account both source and target information.
Moreover, a significant advantage of this approach is that it removes
the dependency on the intermediate segmentation step, which is an
additional component that needs to be trained, as well as being a
source of cascaded errors.

A naive approach to SegFree translation consists in using a sliding
window over the source and target streams, which is moved/updated
following a fixed schedule. For instance, every time a new source word is
received, it is added to the source window and the oldest source word
is discarded. However, \citet{iranzo-sanchez-etal-2021-stream-level}
shows that the target-to-source ratio of the words generated by an MT
system is not constant during the translation of a source stream, and
a system using this approach will end up with source and target
windows whose content is out of sync, as a result of having over or
under-estimated the writing rate of the system.

Our proposed SegFree system solves this issue by replacing the fixed
update schedule of the sliding windows by a memory mechanism. This
mechanism keeps track of which parts of the source stream have already
been translated, along with the associated translations. In addition,
it manages the streaming history by discarding the oldest, already
translated words of the stream, which can be forgotten without
affecting the current translation. As a result, each of the sliding
windows of the SegFree model contains two disjoint chunks: a chunk of
history words, which has been fully processed, and can therefore be
discarded in the future, and the active chunk that needs to be
translated. Once the maximum capacity of the streaming history has
been reached, the oldest source part and its corresponding translation
are discarded. The proposed memory mechanism uses a probabilistic
model in order to decide which part of the source stream has already
been translated and should be moved to the streaming history together
with its translation.

\begin{figure*}
\centering
\begin{tikzpicture}
	\draw(1,1) node[rectangle]{\small $y_i$ = ``duda''};
	\draw(-6,1) node[rectangle]{$t=i$};
	\draw(-6,0) node[rectangle,draw,anchor=south west, fill=gray!40!white, fill opacity=0.4, text opacity=1, label=left:{$\mathcal{X}$}](t1srcold){ \phantom{HH} reduce harmful emissions};
	\draw(-6,-1) node[rectangle,draw,anchor=south west, fill=gray!40!white, fill opacity=0.4, text opacity=1, label=left:{$\mathcal{Y}$}](t1srcold){reducir emisiones nocivas [SEP]};

\draw(0,0) node[rectangle,draw,anchor=south west](t1src){there is no doubt about that the question is how to do};
\draw(0,-1) node[rectangle,draw,anchor=south west](t1tgt){no me cabe ninguna duda};

        \draw(-6.75,-1.5) -> (9.25,-1.5);

	\draw(-3.75,-2) node[rectangle]{$t=i+1$ ~ (Before history update) };
	\draw(-6,-3) node[rectangle,draw,anchor=south west, fill=gray!40!white, fill opacity=0.4, text opacity=1, label=left:{$\mathcal{X}$}](t2srcold){ \phantom{HH} reduce harmful emissions};
	\draw(-6,-4) node[rectangle,draw,anchor=south west,  fill=gray!40!white, fill opacity=0.4, text opacity=1, label=left:{$\mathcal{Y}$}](t2srcold){ reducir emisiones nocivas [SEP]};

	\draw(2.25,-2) node[rectangle]{\small $y_{i+1}$ = ``[SEP]'' ~~~ $x_{\hat a}$ = ``that''};
	\draw(-0.6,-3) node[rectangle,draw,anchor=south west,  fill=black!70!gray, fill opacity=0.4, text opacity=1](t2src){there is no doubt about that \phantom{HH}};
	\draw(4.85,-3) node[rectangle,draw,anchor=south west](t2src){the question is how to do it};
	\draw(-0.6,-4) node[rectangle,draw,anchor=south west,  fill=black!70!gray, fill opacity=0.4, text opacity=1](t2tgt){no me cabe ninguna duda \small{[SEP]}};

        \draw(-6.75,-4.5) -> (9.25,-4.5);
	\draw(-3.75,-5) node[rectangle]{$t=i+1$ ~ (After history update) };

	\draw(1.25,-5) node[rectangle]{\small $y_{i+1}$ = ``[SEP]''};
	\draw(-6,-6) node[rectangle,draw,anchor=south west, fill=black!70!gray, fill opacity=0.4, text opacity=1, label=left:{$\mathcal{X}$}](t2src){there is no doubt about that \phantom{HHH}};
	\draw(-0.25,-6) node[rectangle,draw,anchor=south west](t2src){the question is how to do it};
	\draw(-6,-7) node[rectangle,draw,anchor=south west, fill=black!70!gray, fill opacity=0.4, text opacity=1, label=left:{$\mathcal{Y}$}](t2tgt){no me cabe ninguna duda \small{[SEP]}};

\end{tikzpicture}
\caption{This figure illustrates the memory mechanism in three
  consecutive steps shown by rows. The chunks with shaded color belong
  to the streaming history, while the unshaded chunks are the current
  active source (top) and target (bottom) streams. In the first row
  ($t=i$), the MT system has just generated the last target word $y_i$
  = ``duda''.  In the second row ($t=i+1$), the translation model
  generates the "[SEP]" token, which indicates the end of a target
  segment. At that point, the memory mechanism is activated and
  decides $\hat a=j+5$ with $x_{\hat a}$ = ``that'', so "there is no
  doubt about that" is moved to the streaming history along with the
  current translation.  In the third row, the translation continues,
  but the streaming history has grown too large, so the memory
  mechanism discards the oldest chunk.
	\label{memory_mechanism}}

\end{figure*}
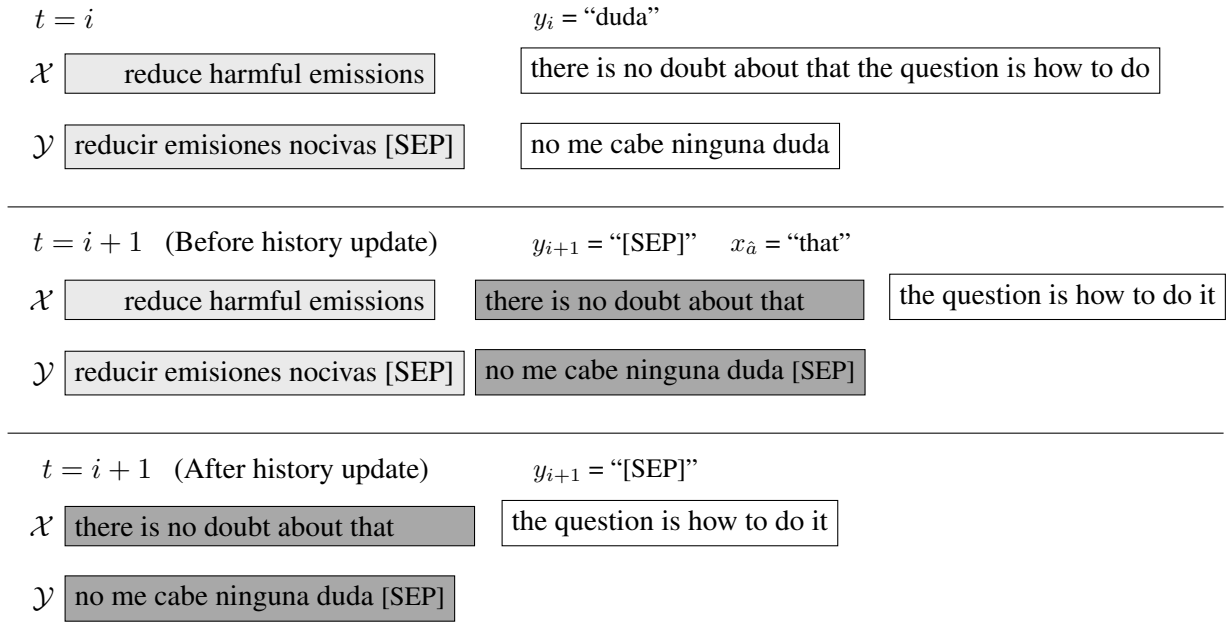

Formally, let $\mathcal{X}=\{x_1, x_2, \dots, x_J\}$ be the source
stream and $\mathcal{Y}=\{y_1, y_2, \dots, y_I\}$, the target stream.
Let $x_j^{j\sp{\prime}}$ be a chunk of active source words and $\hat
y_i^{i\sp{\prime}}$ a partial translation of that chunk.  Every time
the translation model generates the end-of-segment token ("[SEP]"),
the memory mechanism is invoked and selects a source position $\hat a
\in [j,{j\sp{\prime}}]$ based on the current status of the
translation.  After this decision, $x_j^{\hat a }$ and $\hat
y_i^{i\sp{\prime}}$ will be moved to the streaming history, and the
translation will continue with $x_{\hat a + 1}^{j\sp{\prime}}$. This
is graphically shown in Figure~\ref{memory_mechanism}.

For this work, the memory mechanism takes the current active source
and target chunks, $\bm{x}=x_j^{j\sp{\prime}}$ and $\bm{\hat
  y}=y_i^{i\sp{\prime}}$ respectively, and uses a log-linear model
composed by a series of feature functions $h_f(a,\bm{x},\bm{\hat y)}$
which provide a score for every position $a$ of the active source
chunk. The probability of position $a$ being the last position that
has been translated is estimated as
\begin{equation}
	p(a|\bm{x},\bm{\hat y} ) = \frac{\prod_f h_f(a,\bm{x},\bm{\hat y})^{\lambda_f}}{\sum_{a'} \prod_f h_f(a',\bm{x},\bm{\hat y})^{\lambda_f}}. \label{eq:1}
\end{equation}

After applying the log-transformation, the position of the last source
word $\hat a$ to be moved to the streaming history is chosen as
\begin{equation} 
	\hat a = \argmax_a \sum_f \lambda_f \log h_f(a,\bm{x},\bm{\hat y}).
\end{equation}

The weights of the feature functions, $\bm{\lambda} \in \mathbb{R}^F$,
are optimized using gradient descent by minimizing the conventional
cross-entropy loss over a set of samples. Specifically, each sentence
pair can be understood as a classification sample for a task with
classes $C=\{1, 2, \dots, |\bm{x}|\}$ and correct label $\hat
C=|\bm{x}|$. The abbreviated pseudocode for SegFree inference is
presented in Figure~\ref{alg:inference}.

\begin{figure}[!htb]
\small
	\begin{minted}{python}
def memory_mechanism(states):
  # Compute unnormalized p(a|x,y)
  # for all a (numerator of Eq. 1)
  scores = score(states["src"][-1],
                 states["tgt"][-1])
  a_hat = argmax(scores)

  # Translated part, sent to history
  trans = states["src"][-1][:a_hat]
  # Untranslated part, to be kept
  rest = states["src"][-1][a_hat:]

  states["src"][-1] = trans
  states["src"].append(rest)
  states["tgt"].append([])

  # Remove src/tgt pairs until
  # memory buffer is below max len
  filter_to_max_len(states)

states = {"src": [[]], "tgt": [[]]}
while True:
  action, word = policy(states)
  # Normal READ action
  if action == READ:
    states["src"][-1].append(read())
  # Normal WRITE action
  elif word != "[SEP]":
    states["tgt"][-1].append(word)
    if word == "[END]":
      break
  # Only when writing "[SEP]"
  else:
    memory_mechanism(states)
	\end{minted}
       \caption{Python-like pseudocode for our proposed
          SegFree model with memory mechanism. \label{alg:inference}}
\end{figure}

The SegFree model presented in this work uses a memory mechanism based
on the following feature functions\footnote{Note that in these
definitions, $a$ is the relative position from the start of the
active chunk, that is, $a\in[1, |\bm{x}|]$.}:
\begin{itemize}
	\item A reverse translation model (\textit{Reverse MT}). The
          score for each source position is given by a reverse
          translation model that computes the probability of the
          partial translation $x_j^{a}$ followed by the end-of-sentence symbol
		("</s>"). That is,
          $h_f(a,\bm{x},\bm{\hat y})=p_{y \rightarrow
		x}([x_j^{a}, \textrm{</s>}]|\bm{\hat y})$,
		computed as the product of token-level probabilities including $\textrm{</s>}$.
		For this work, our
		reverse model uses the same architecture and training data as the forward
		model, but the translation direction has been switched during training.
	\item A normal distribution conditioned by a linear regression (LinReg)
          model $h_f(a,\bm{x},\bm{\hat y})=\mathcal{N}(a \mid
          \theta_{\mu} \cdot |\bm{\hat y}|,\theta_{\sigma}^2)$,
          estimated with Ordinary Least Squares.
\end{itemize}

Apart from the two aforementioned features, we also tried some other
approaches that were not able to improve the results, either in
isolation or in combination with other features. Specifically, we
tried predicting both the mean and the variance rather than only the
mean with the linear model as well as replacing the linear models with
higher order models. Furthermore, we also collected counts of source
lengths for every different target length and used them to estimate
the segmentation probability, as well as using the source length,
target length and length ratios as features. Lastly, we also tried a
neural-based regressor that predicts the mean and variance.

Additionally, we also compare our log-linear approach with the naive
SegFree approach (\textit{Naive}), which uses a sliding window with a
fixed offset $r$. The median target-to-source length ratio is used as
the offset. Then, during inference time, whenever the ``[SEP]'' token
is emitted, the source sliding window is moved a fixed number of
positions based on this pre-computed ratio. Specifically, $\hat a =
\max(\lfloor |\bm{y}|/{r} \rfloor, 0).$

\section{Experimental settings}
\label{sec:settings}

\subsection{Datasets}

Results are presented for four translation directions: English to
German, Spanish and French, and German to English.  The models
involving German were trained using datasets available for the IWSLT
2022 shared tasks~\cite{anastasopoulos-etal-2022-findings}.  The
specific datasets are reported in Table~\ref{tab:datasets}.  The other
systems were trained using open data from OPUS
NLPL~\cite{Tiedemann09}.  Specifically, the English to French and
Spanish models were trained using 305M and 327M sentence pairs,
respectively.  All sentences for which document-level information is
present are augmented with their corresponding streaming
history~\cite{iranzo-sanchez-etal-2022-simultaneous}, by concatenating
the previous source and target sentences until a maximum length of 50
words has been reached.

\begin{table}
	\centering
\caption{Overview of the datasets used for training, including number
  of sentence pairs (in thousands) and the availability of
  document boundaries, which used for constructing samples with
  streaming history. Note that Europarl~\cite{koehn-2005-europarl} is
  excluded from the training data, in order to avoid overlap with
  Europarl-ST. \label{tab:datasets}}
\begin{tabular}{lrc}
	Corpus & \# sentences (K) & Doc \\ \hline
	News-comm. v16 & 398 & \checkmark\\
	Tilde-Rapid & 1531 & \checkmark\\
	MuST-C & 250 & \checkmark \\
	Europarl-ST & 45 & \checkmark \\
	ParaCrawl & 82638 & - \\
	CommonCrawl & 2399 & -\\
	WikiTitles & 1474 & -\\
	WikiMatrix & 6227 & -\\
	LibriVox & 51 & -\\
\end{tabular}
\end{table}

In order to enable simultaneous translation, the prefix-training data
augmentation technique~\cite{arivazhagan-etal-2020-translation} is
used.  One partial translation pair is generated for each sentence
pair in the original corpus (which already includes the streaming
history) by randomly selecting a partial prefix of both, source and
target sentences. If a given sentence pair contains streaming history
information, the streaming history is left unchanged, and prefix
generation is only applied to the current sentence pair.  The model is
trained on the concatenation of both, the original training data and
the partial prefix data.  Figure~\ref{fig:corpus_ex} shows a graphical
overview of how the training data was constructed.

\begin{figure*}

\centering

\begin{tabular}{ll}
Original & Prefix-augmented \\
\hline \hline
I'm going to talk today about energy and climate.  & I'm going to talk today \\ 
Heute spreche ich zu Ihnen über Energie und Klima. & Heute spreche ich zu Ihnen \\
\\
{\color{black!45}Think about it. [SEP]} The PC is a miracle. & {\color{black!45}Think about it. [SEP]} The PC is \\ 
{\color{black!45}Denk darüber nach. [SEP]} Der PC ist ein Wunder. & {\color{black!45}Denk darüber nach. [SEP]} Der PC ist \\
\end{tabular}
 \caption{Illustrated example of how the training set was prepared.
   One prefix training version is generated for each sentence pair by
   discarding a portion of both the source and target sentences.  The
   first row shows a source-target sample without streaming history
   that is randomly prefixed.  The second row is a source-target
   sample including streaming history (shown in light gray), in which
   prefix augmentation is only applied to the current sentence to be
   translated, but the history remains unchanged.  The final dataset
   contains both, the \textit{Original} and the
   \textit{Prefix-augmented} samples, so the size of the training set
   is doubled. \label{fig:corpus_ex}}
\end{figure*}

The source side of the dataset is lowercased and punctuation marks are
removed in order to simulate the output of a streaming ASR
system. SentencePiece~\cite{kudo-richardson-2018-sentencepiece} is
used to learn 50k subword units. The SentencePiece whitespace meta
symbol "\_" is used as a suffix instead of a prefix, so that a full
word can be written once its last subword has been written, without
having to wait for the model to generate the next subword.  Both
Segmented and SegFree systems are trained exactly with the same data,
with the only difference between the two being that the training data
of the SegFree model has been processed to remove ``[SEP]'' tokens
from the source side of the data, in order to mimic the inference
condition in which no end-of-chunk information is available.  The
baseline segmenters and the SegFree feature functions were trained
with the MuST-C v2 train set. The feature function weights were then
optimized with the samples of the MuST-C v2 dev set.

\subsection{Translation models}

Both Segmented and SegFree systems use a Transformer BIG
model~\cite{Vaswani2017}, trained following the streaming-history
setup of \citet{iranzo-sanchez-etal-2022-simultaneous}.  We opted to
use a conventional Transformer trained with prefix-augmented
data~\cite{arivazhagan-etal-2020-translation} rather than their masked
wait-$k$~\cite{ma-etal-2019-stacl,elbayad20} training as the results
of \citet{arivazhagan-etal-2020-translation} show that is a better
choice.  No specific architecture changes are applied for the
simultaneous task, as the model learns to generate simultaneous
translations thanks to the data augmentation regime. At inference
time, the latency of the models is controlled with a wait-$k$ policy~
\cite{ma-etal-2019-stacl}. The words in the streaming history are
ignored for the purposes of the policy, that is, only the words in the
active chunk are taken into account when deciding between a READ or a
WRITE operation. Speculative Beam
Search~\cite{zheng-etal-2019-speculative} with a beam size of 4 is
used to generate hypotheses. The best scoring hypothesis is selected,
and then only the amount of words indicated by the wait-$k$ policy
will be committed as a WRITE operation, the rest are discarded. The
search is always initialized with a target prefix consisting of the
already committed target words. Every time a target sentence is
committed (indicated by the ``[SEP]'' token), the length of the
streaming history is checked, and if the maximum history size is
exceed in either the source or the target side, pairs of segments are
removed from the streaming history until the maximum word length (50)
is no longer exceeded.

Apart from the aforementioned Segmented and SegFree systems, a system
following the approach of \citet{sukanta2022} has also been trained to
serve as an additional baseline. This system uses the same training
data as the other systems, but rather than sentence-based samples, the
data is first aligned at the word level using
\textit{fastAlign}~\cite{dyer-etal-2013-simple}, and then the
algorithm proposed by \citet{sukanta2022} is used to extract window
pairs for training.  The results for this system are reported as
\textit{Window Retrans}.

\subsection{Segmented setting}

For the Segmented setting, the Direct Segmentation (DS) approach
described
in~\cite{iranzo-sanchez-etal-2020-direct,iranzo-sanchez-etal-2022-simultaneous}
is used, which is a streaming segmenter with a small future window.
The DS approach considers the segmentation as a classification problem
and decides, for each source word, whether it is the end of a chunk or
not. The detected chunks are then translated by the MT system. The
end-of-chunk events detected by the segmenter are conveyed to the MT
system by inserting the ``[SEP]'' token into the source text received
by the MT system.

The original DS system used an RNN-based classifier, however our
experiments revealed that replacing the RNN-based model with a
finetuned XLM-RoBERTa model~\cite{conneau-etal-2020-unsupervised}
provides a significant translation quality gain.  Indeed, a Large
XLM-RoBERTa model was selected as it outperformed both, the original
RNN segmenter and the Base XLM-RoBERTa version, providing an even
stronger segmented baseline.  DS models were trained with history size
10 and a different system was trained for each value of future window
$w\in\{0,1,2,4\}$.  The results of an Oracle segmenter (DS-Oracle)
using the reference source sentence segmentation are also reported as
an upper bound to better understand the effect of the segmentation.

\section{Results}
\label{sec:results}
SegFree and DS-based models follow a wait-$k$ translation policy.  We
report 10 results for each system, one for each $k\in[1,10]$, in order
to explore the latency-quality tradeoff.  Each video belonging to the
evaluation set is translated independently from the other videos in
the set. Because both, the DS and the SegFree approach, create their
own segmentation that does not match the reference one, the hypotheses
are re-aligned with the reference translation using minimum edit
distance~\cite{matusov-etal-2005-evaluating} before computing the
quality measure BLEU~\cite{papineni-etal-2002-bleu}\footnote{BLEU|nrefs:1|case:mixed|eff:no|tok:13a|smooth:exp|version:2.2.1}. Likewise,
stream-level latency~\cite{iranzo-sanchez-etal-2021-stream-level} is
computed using minimum edit distance so that both approaches can be
compared.  The average of the Average Lagging
(AL)~\cite{ma-etal-2019-stacl} value of each individual video is
reported.

The quality-latency tradeoff of the WindowRetrans approach is
controlled using two hyperparameters: $w$, which is the size of the
window that is re-translated at each step, and $r$, the match
threshold that needs to be reached by a hypothesis to be considered a
match. Similarly to \citet{sukanta2022}, we test $w\in\{8,12,16,20\}$
and $r\in\{0.1,0.2,\dots,0.7\}$.  WindowRetrans does have flickering,
unlike the other systems.  We follow the conventional practice of
evaluating on the final
text~\cite{ArivazhaganCTMB20,arivazhagan-etal-2020-translation,yao-haddow-2020-dynamic}
and disregard any flickering for quality evaluation. This enables us
to compare the WindowRetrans system with the other proposed systems,
assuming an ideal situation in which flickering can be safely
discarded.  However, in practice we would not be able to do this, as
the output of the downstream TTS system cannot be changed once it has
been received by the listener.

\begin{figure}[htb] 
\centering
\includegraphics[width=.49\textwidth]{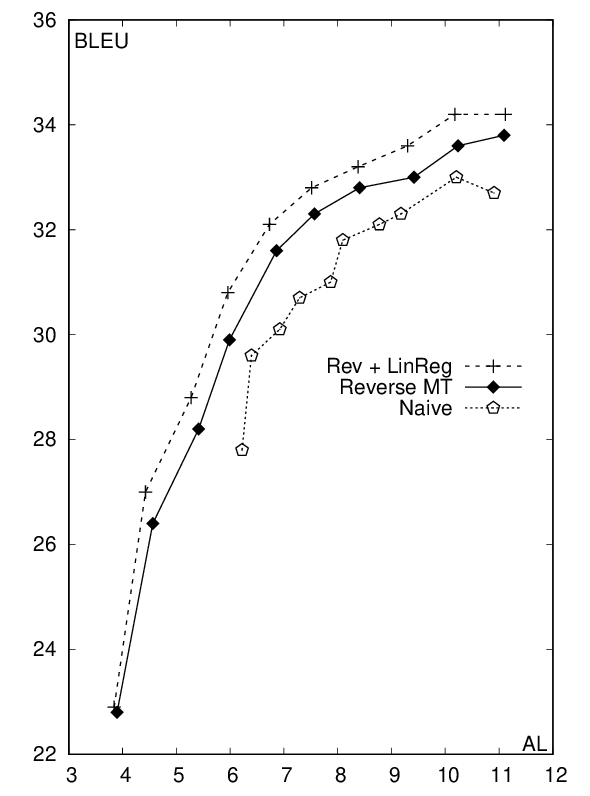}
	\caption{Comparison of BLEU vs. AL between the Naive SegFree
          system and SegFree systems based on two different setups of
          feature functions (Reverse-MT and Reverse-MT + LinReg) on
          the English to German Europarl-ST dev set. There are 10 results
	  for each system, one for each $k\in[1,10]$.
\label{fig:feature_experiments}}
\end{figure}

Figure~\ref{fig:feature_experiments} shows BLEU vs. AL of SegFree
systems when using two different combinations of feature functions
(Reverse-MT and Reverse-MT + LinReg) compared with the Naive SegFree
system, evaluated on the English to German Europarl-ST dev set.
Unsurprisingly, the Naive approach underperforms the other two
systems.  The use of a fixed offset in the Naive approach is a
limiting factor for the translation quality, as both the source and
target streams are assumed to progress at the same rate, irregardless
of their actual content. Every time a target sentence is produced, a
fixed number of source words are considered to have been
translated. This means that on some occasions the actual writing rate
may be underestimated, and on other occasions it may be overestimated.
In this case, the results suggest that the writing rate might have
been underestimated, which in turn causes high latency even for low
values of $k$.  In contrast, our proposed SegFree system with
Reverse-MT feature works significantly better than the Naive baseline,
because it can dynamically update the streaming history based on the
source and target streams, instead of being constrained by a fixed
rate. Thus, if a source chunk containing many high fertility words is
translated, the system can take this into account when updating the
streaming history. This avoids the problem of marking untranslated
words as already translated, which is what would have happened in the
Naive approach. On top of this, combining the Reverse-MT feature with
the Linear Regression feature (Reverse-MT + LinReg) further improves
the results, as the Linear Regression feature smooths the
probabilities given by the Reverse-MT model. Based on this result, the
Reverse-MT + LinReg system is selected for further experimentation.

\begin{figure}[htb] 
\centering
\includegraphics[width=.51\textwidth]{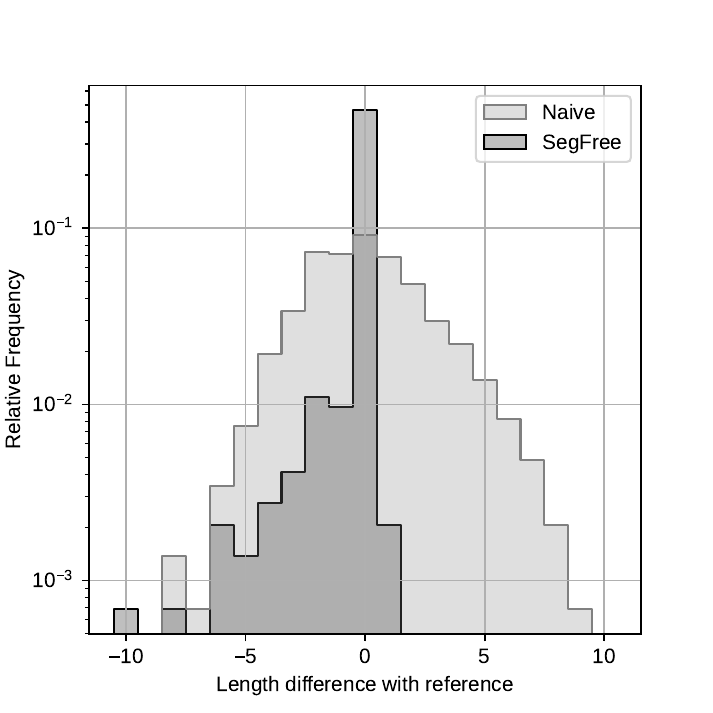}
	\caption{Logarithmic scale distribution of the difference
          between the source chunk length computed by the Naive and
          Reverse-MT SegFree models and the reference length used by
          the DS-Oracle model. Results computed on the English to
          German Europarl-ST dev.  Negative numbers indicate that the
          length was underestimated by the memory mechanism, whereas
          positive numbers indicate that it was overestimated.
\label{fig:analysis}}
\end{figure}

We test the previous hypothesis by taking the translations generated
with the DS-Oracle and feeding the memory mechanism of the Naive and
Reverse-MT models with the appropriate source context. Because the
DS-Oracle tells us which source words have actually been used to
generate the translation, we can test if the memory mechanism is able
to correctly identify these words. Figure~\ref{fig:analysis} shows the
difference in length between the hypothesis generated by the memory
mechanism and the DS-Oracle.  It can be observed how the SegFree
system is very good at detecting the correct position, except for some
cases in where the length is underestimated.  In contrast, the Naive
approach cannot adapt its prediction depending on the content of the
actual translation, and as a result it performs significantly worse at
selecting the right position to update the source stream.

\begin{figure}[htb] 
\centering
\includegraphics[width=.49\textwidth]{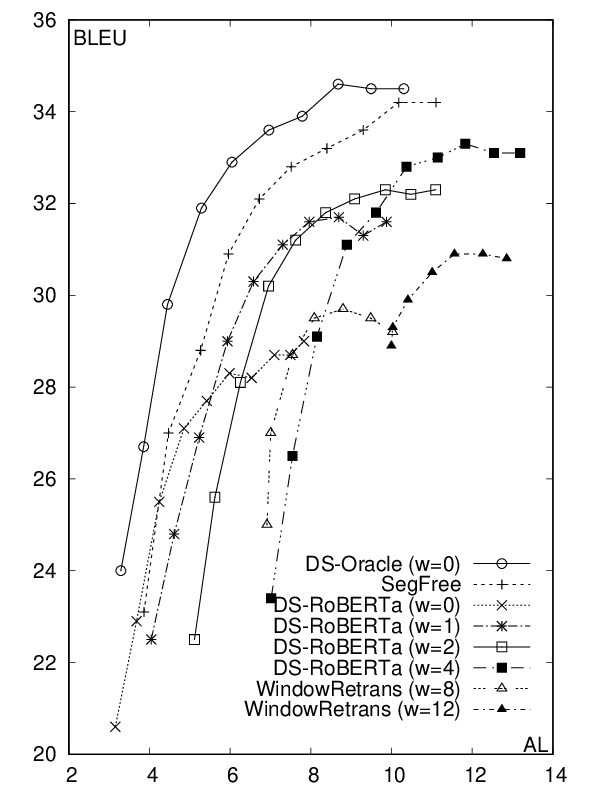}
\caption{Comparison of BLEU vs. AL between the proposed SegFree
  approach and the baseline models on the English to German Europarl-ST dev
  set. For the WindowRetrans models,
  each point corresponds to a different $r\in\{0.1,0.2,\dots,0.7\}$.\label{fig:Europarl_ST_dev}}
\end{figure}

Figure~\ref{fig:Europarl_ST_dev} shows a comparison between the
SegFree approach and the selected baselines on the English to German
Europarl-ST dev set. The DS-RoBERTa quality/latency trade-off is very
dependant on the size of the future window $w\in\{0,1,2,4\}$.  It can
be observed how $w=0$ only remains competitive for low latencies, but
it quickly plateaus between 28 and 29 BLEU points. The lack of a
future window means that the segmentation decisions are less informed,
and the translation quality does not greatly increase even if the
translation model is given more context.  Moving from $w=0$ to $w=1$
provides a significant quality boost, and the model is able to reach
31.7 BLEU points.  Larger future window values ($w=2$ and $w=4$)
provide further quality improvements, reaching a maximum of 32.3 and
33.3 BLEU points, respectively, but the additional latency introduced
by the segmenter does not make them competitive choices. This is
consistent with the results of other works that use the DS
segmenter~\cite{IRANZOSANCHEZ2021}. Once the DS-RoBERTa model has one
or two future context tokens, it is better to allocate additional
latency to the MT model in order to avoid diminishing returns. Based
on this, $w=1$ was selected for the final evaluation on the test
sets. There is a gap of around 3 BLEU points between the DS-RoBERTa
systems and the DS-Oracle across all latency regimes. This gap
illustrates the loss of performance incurred when using an imperfect
segmentation, as well as the upper bound of performance that could be
achieved using a perfect segmenter.

The proposed SegFree system clearly outperforms the DS-RoBERTa systems
at mid and high latencies, and performs similarly to the best
DS-RoBERTa system at low latencies. The SegFree system achieves this
quality improvement by having access to the original source stream and
letting the MT system take the decision where the segment delimiter
should be placed. Moreover, the SegFree system achieves these results
consistently, whereas the DS approach needs multiple segmenters with
different $w$ in order to stay competitive. This highlights another
advantage of moving beyond a segmenter system, as the latency of the
translation only depends on the policy of the MT system. The
WindowRetrans system is far behind the performance of both the SegFree
and the best DS-RoBERTa system. The results for WindowRetrans with
$w=16$ and $w=20$ are not included in Figure~\ref{fig:Europarl_ST_dev}
as they had even worse latency-quality trade-off. The configuration
with $w=8$ was selected for further evaluation.

\begin{figure*}[htb] 
	\centering
	\includegraphics[width=.355\textwidth]{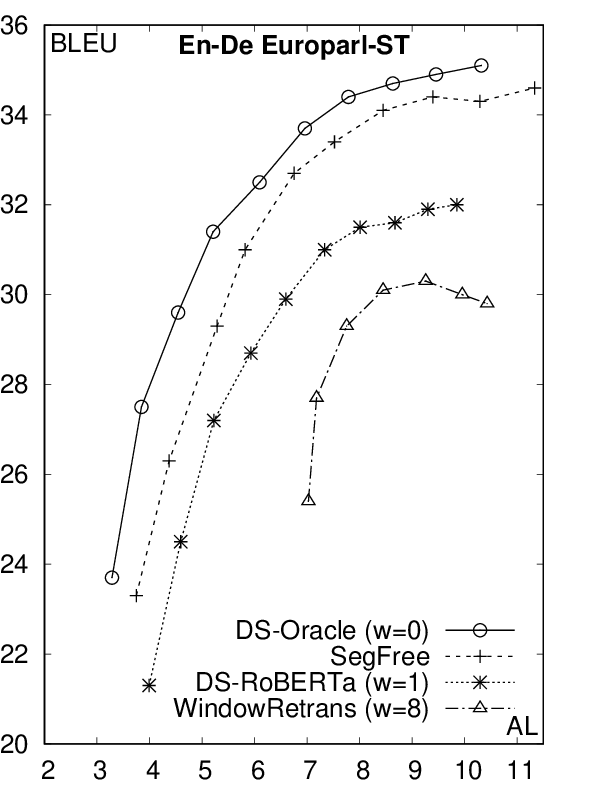}\hspace*{-5mm}
	\includegraphics[width=.355\textwidth]{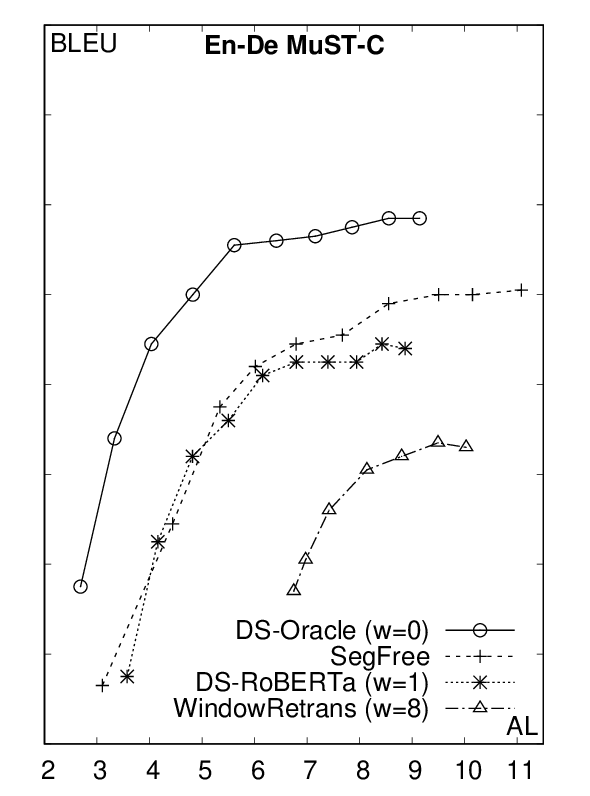}\hspace*{-5mm}
	\includegraphics[width=.355\textwidth]{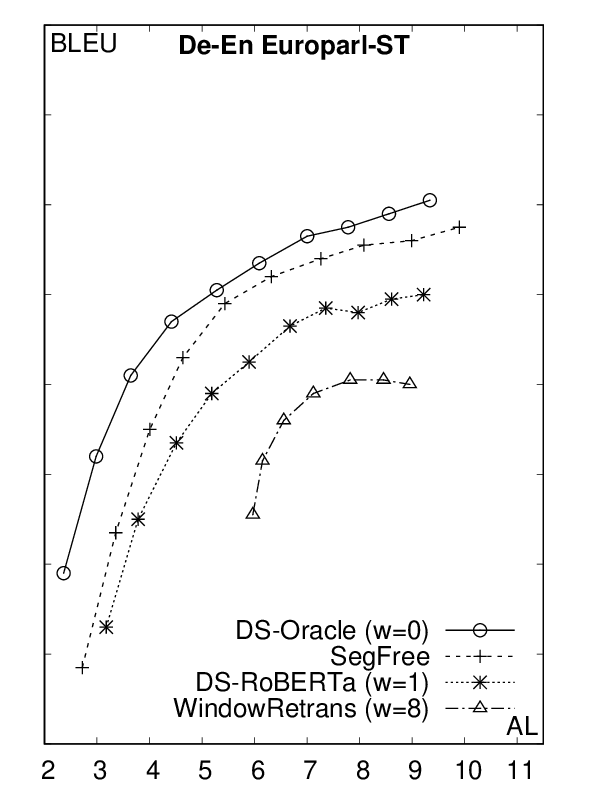}
	\caption{BLEU vs. AL on the English to German Europarl-ST
          (left) and MuST-C (center) test sets, and on the German to
          English Europarl-ST (right) test
          set.\label{fig:test_combined}}
\end{figure*}

After performing hyperparameter exploration on the Europarl-ST dev
set, the DS-RoBERTa, DS-Oracle, WindowRetrans and SegFree systems were
evaluated on the selected test sets.  Figure~\ref{fig:test_combined}
reports BLEU vs. AL results, from left to right, on the English to
German Europarl-ST and MuST-C test sets, and the German to English
Europarl-ST test set.  Statistical significance tests using bootstrap
resampling~\cite{koehn-2004-statistical,post-2018-call} were conducted
to test whether differences between systems were significant, with
1000 bootstrap resamples per test. Each system was compared with each other
system within $\pm 0.3$ AL.

For the English to German Europarl-ST test set, the SegFree
outperforms the DS-RoBERTa system by a wide margin. There is a gap of
around 2 BLEU points across all latency regimes, and this gap grows up
to 2.6 BLEU when comparing the best results (34.6 BLEU points for
SegFree and 32.0 BLEU points for DS-RoBERTa).  The SegFree is 1.1 BLEU
points behind the DS-Oracle system at medium latencies (AL$\simeq$5),
and this difference decreases at higher latencies (0.5 BLEU points for
AL$\simeq$9). BLEU differences across systems were statistically
significant.

On the MuST-C test set, both, the SegFree and the DS-RoBERTa systems
perform similarly at low and medium latencies. The SegFree system does
significantly outperform the DS-RoBERTa system for AL$\geq$7.6,
reaching a maximum of 30.1 BLEU points, whereas the DS-RoBERTa system
provides 28.8 BLEU points. The DS-Oracle is significantly better than
both the SegFree and DS systems.

Lastly, the German to English Europarl-ST test set results show that
the SegFree system significantly outperforms the DS-RoBERTa system
across all latency regimes. For example, there is a gap of 1.9 BLEU
points for AL$\simeq$ 4.5, and a gap of 1.1 BLEU points for AL$\simeq$
8.0.  When comparing the DS-Oracle and the SegFree system, the
DS-Oracle is not significantly better for $k\in\{5,6,7\}$.

The WindowRetrans system shows a similar trend to the one that was
observed on the dev set. As the value of $r$ is increased, so does the
quality of the translation and the latency. The quality plateaus when
$r=0.5$ or $r=0.6$ is reached, and further increases on $r$ tend to
degrade the performance. The difference in quality between this
approach and the DS-RoBERTa model is statistically
significant. Figure~\ref{fig:test_combined_BLEURT} reports the results
of the different translation systems when evaluated with the
BLEURT-20~\cite{pu-etal-2021-learning} neural measure. We observe no
significant differences when compared with the evaluation carried out
using BLEU.

Figure~\ref{fig:test_combined_enfr} reports the results for the
English to French system, evaluated on the Europarl-ST and MuST-C test
sets. The results show a similar pattern on both test sets: The
DS-RoBERTa system outperforms the WindowRetrans baseline across all
latency ranges, and is in turn surpassed by the proposed SegFree
system. A small gap remains between the SegFree system and the
DS-Oracle, and this gap is not statistically significant at some
latency conditions. For AL $\simeq 9$, there is a gap of 0.4 BLEU
between the DS-Oracle and the SegFree system on the Europarl-ST test,
and a gap of 3.2 BLEU between the SegFree system and the DS-RoBERTa
system. BLEU scores for WindowRetrans were too low (32.8) and omitted
in Figure~\ref{fig:test_combined_enfr} for the sake of clarity.  For
the MuST-C test, there is a larger gap of 1.7 BLEU between the
DS-Oracle and the SegFree system, and a gap of 2.9 BLEU between the
SegFree and the DS-RoBERTa system.

Next, Figure~\ref{fig:test_combined_enes} reports the results for the
English to Spanish system.  The results follow a similar trend to
previous test sets. For the Europarl-ST test (AL $\simeq 8$), there is
a gap of of 1.7 BLEU between DS-Oracle and SegFree systems, and a gap
of 4.0 between the SegFree and the DS-RoBERTa systems. The
WindowRetrans result is 1.2 BLEU lower than that of DS-RoBERTa.  For
the MuST-C test, these gaps are 1.7, 2.6 and 2.8 BLEU, respectively.

Figures~\ref{fig:test_combined_enfr_BLEURT}
and~\ref{fig:test_combined_enes_BLEURT} report BLEURT curves rather
than BLEU for the same datasets and languages pairs.  The English to
French BLEURT results are shown in
Figure~\ref{fig:test_combined_enfr_BLEURT}, whereas the English to
Spanish results are shown in
Figure~\ref{fig:test_combined_enes_BLEURT}.  As in previous cases,
there are no relevant changes regarding system ordering or gaps
between systems. Similar conclusions are reached when evaluating with
either of the measures, BLEU or BLEURT.

\subsection{Computational efficiency}

Both the DS-RoBERTa and the SegFree systems have one additional neural
model than the Naive baseline. Both are Transformer-based models with
different architectures, but a similar number of parameters (300M). We
collect results from all of our experiments, carried out on a machine
with a i9-10920X CPU and an NVIDIA 3090 GPU.  The cost of running this
additional neural model once is on average 15ms $\pm$ 2ms (min. 10ms,
max. 35ms) for the reverse model integrated into the SegFree memory,
and 19ms $\pm$ 1ms (min. 15ms, max. 50ms) for the DS-RoBERTa
system. The DS-RoBERTa system is called every time a new source word
is read, whereas the SegFree reverse model model is only called when
the "[SEP]" token is generated by the translation model.  For all
intents and purposes, both approaches can be assumed to have the same
computational cost.

\begin{figure*}[htb]
        \centering
        \includegraphics[width=.355\textwidth]{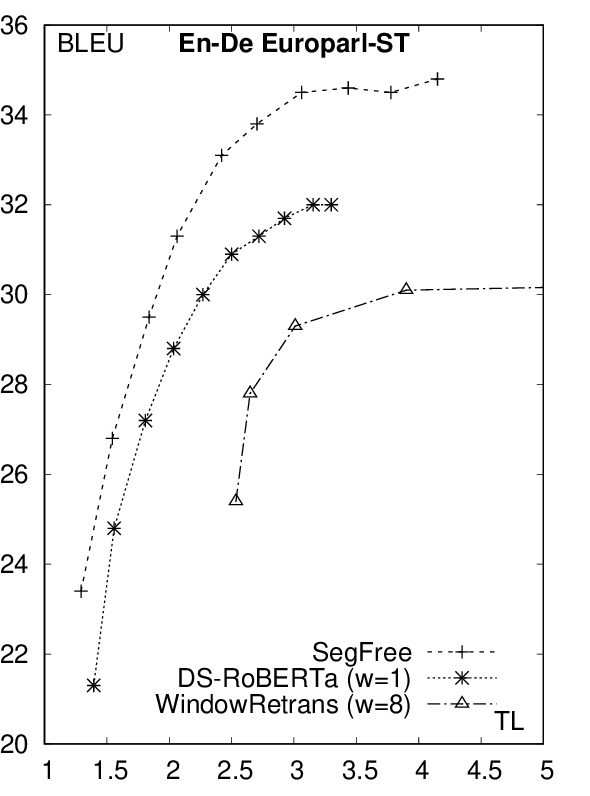}\hspace*{-5mm}
        \includegraphics[width=.355\textwidth]{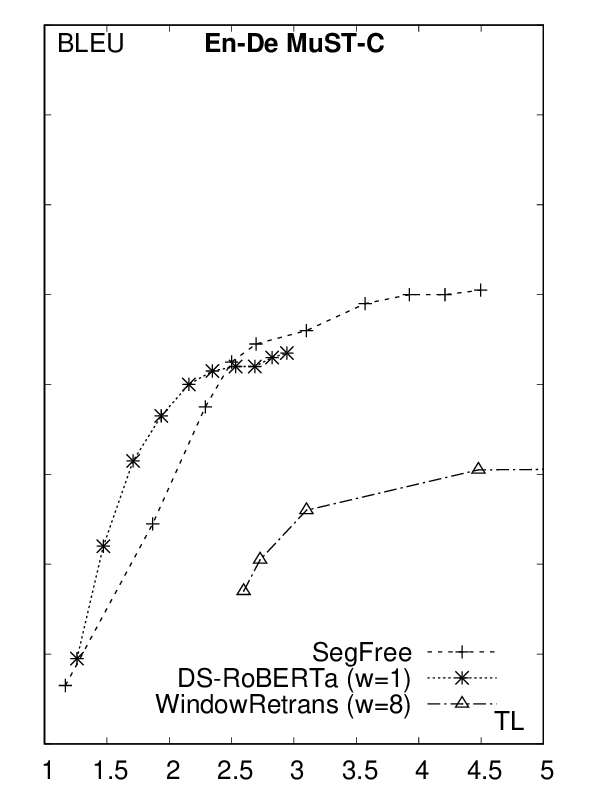}\hspace*{-5mm}
        \includegraphics[width=.355\textwidth]{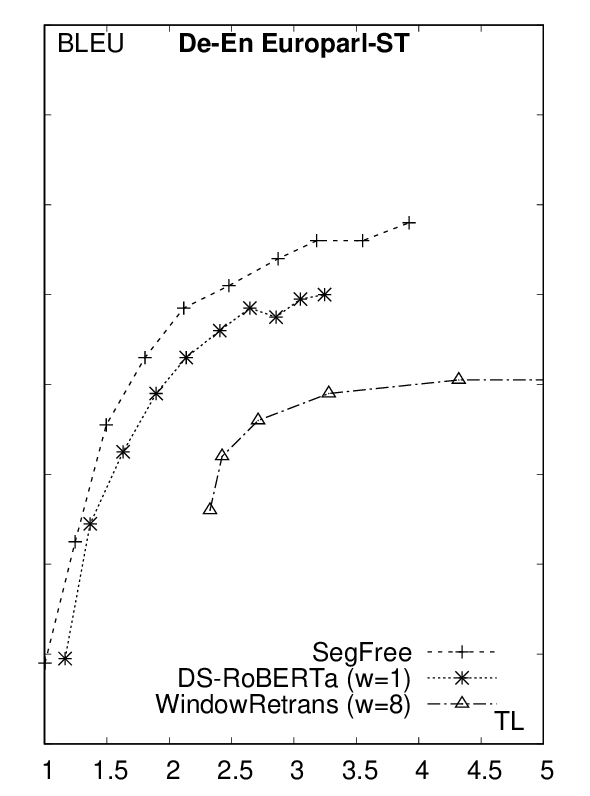}
        \caption{BLEU vs. TL on the English to German Europarl-ST
          (left) and MuST-C (center) test sets, and on the German to
          English Europarl-ST (right) test
          set.\label{fig:test_combined_tl}}
\end{figure*}

%The results from the previous section are reported with measures that
%are not computationally aware.
Figure~\ref{fig:test_combined_tl} reports the results using the
computationally-aware Translation Lag~\cite{ArivazhaganCTMB20} measure
to check if there are any relevant differences with the stream-level
AL results. In order to obtain timestamps for the words on the source
side, we forced aligned the transcriptions with the audio using an
off-the-shelf ASR system.

The results for both the English to German and German to English
Europarl-ST test sets are similar for either AL or TL. On the MuST-C
test set, there is a region on the low-latency regime where the
DS-RoBERTa system performs better than that of SegFree.  The
computational cost of both models is the same, so this gap reveals a
difference in behaviour in the translation of certain words. AL
assumes a constant cost for every word, whereas in TL the cost is
estimated based on the source audio timestamp. This means that pauses
and other similar phenomena are accounted with TL, whereas they would
be ignored for latency computation with AL.

For the WindowRetrans system, the results are similar for $r \leq
0.4$. The match threshold $r$ controls the minimum acceptable match
between the translation of the current window and the output
stream. If this match is not reached, the system extends the
translation window by one word and generates another translation until
the minimum match is reached, or five re-translations have already
been generated.  For $r>0.4$, the system is forced to generate too
many re-translations and it starts falling behind the speaker.

\section{Conclusions}
\label{sec:conclusions}

This work introduces a novel SegFree approach to STR-MT, that can
directly translate an unbounded text stream without having to rely on
an intermediate segmenter.  This is achieved by letting the MT system
decide where the segment delimiters are placed, and delaying this
decision until the translation has been generated. In addition, a
memory mechanism keeps track of which parts of the stream have already
been translated, and can therefore be forgotten when needed, and which
parts remain untranslated and must be kept. The SegFree approach
avoids the performance degradation introduced by a segmenter model,
and is able to take into account additional information from both,
source and target streams, when generating the output translation and
segment delimiters. The experiments have shown how the SegFree system
is able to significantly outperform the competing DS-RoBERTa approach
across six of the seven test sets. Furthermore, the SegFree approach
is able to match the performance of the oracle segmenter in the
Europarl-ST German to English test set.  These results validate the
performance of the SegFree approach across multiple domains and
translation directions. More importantly, the SegFree approach
eliminates the need of an intermediate segmenter system in a cascaded
system.  As a result, the SegFree approach is not only better in terms
of quality, but it also lets the MT system retain full control over
the translation policy.

As a future work, the proposed SegFree memory mechanism has been
instantiated with a Reverse-MT feature and a Linear Regression
feature, but the generic formulation allows for any arbitrary feature
function to be used.  Likewise, the SegFree approach has been tested
with static translation policies, but it could also be applied to a
dynamic translation policy. The SegFree approach opens the doors to
further research that moves away from local, sentence level
translation with limited context, into a fully-fledged contextual
translation system augmented with a dynamic history that keeps the
appropriate context.

\section*{Acknowledgments}

We thank our action editor and the anonymous reviewers 
for their helpful and constructive comments on the paper. 
% for their valuable feedback, which significantly enhanced the paper’s quality.
% for their valuable feedback and discussions.
%
The research leading to these results has received funding from
European Union's Horizon 2020 research and innovation program under
grant agreement no. 952215 and
EU4Health Programme 2021-2027 as part of Europe’s Beating Cancer Plan
under Grant Agreements nos. 101056995 and 101129375; 
and from the Government of Spain's
grant PID2021-122443OB-I00 funded by MCIN/AEI/10.13039/501100011033 and by
“ERDF A way of making Europe”,
grant PDC2022-133049-I00 funded by MCIN/AEI/10.13039/501100011033 and by
the “European Union NextGenerationEU/PRTR” and
FPU scholarship FPU18/04135.
The authors gratefully acknowledge the financial support of the Generalitat Valenciana under project IDIFEDER/2021/059.
% “CLUSTERIA: CLÚSTER DE COMPUTACIÓN DE ALTAS PRESTACIONES PARA INTELIGENCIA ARTIFICIAL CONFIABLE”)

\bibliography{tacl2021,anthology}

\begin{thebibliography}{49}
\expandafter\ifx\csname natexlab\endcsname\relax\def\natexlab#1{#1}\fi

\bibitem[{Agrawal et~al.(2018)Agrawal, Turchi, and
  Negri}]{agrawal-etal-2018-contextual}
Ruchit Agrawal, Marco Turchi, and Matteo Negri. 2018.
\newblock \href {https://aclanthology.org/2018.eamt-main.1} {Contextual
  handling in neural machine translation: Look behind, ahead and on both
  sides}.
\newblock In \emph{Proceedings of the 21st Annual Conference of the European
  Association for Machine Translation}, pages 31--40, Alicante, Spain.

\bibitem[{Anastasopoulos et~al.(2022)Anastasopoulos, Barrault, Bentivogli,
  Zanon~Boito, Bojar, Cattoni, Currey, Dinu, Duh, Elbayad, Emmanuel,
  Est{\`e}ve, Federico, Federmann, Gahbiche, Gong, Grundkiewicz, Haddow, Hsu,
  Javorsk{\'y}, Kloudov{\'a}, Lakew, Ma, Mathur, McNamee, Murray,
  N{\v{a}}dejde, Nakamura, Negri, Niehues, Niu, Ortega, Pino, Salesky, Shi,
  Sperber, St{\"u}ker, Sudoh, Turchi, Virkar, Waibel, Wang, and
  Watanabe}]{anastasopoulos-etal-2022-findings}
Antonios Anastasopoulos, Lo{\"\i}c Barrault, Luisa Bentivogli, Marcely
  Zanon~Boito, Ond{\v{r}}ej Bojar, Roldano Cattoni, Anna Currey, Georgiana
  Dinu, Kevin Duh, Maha Elbayad, Clara Emmanuel, Yannick Est{\`e}ve, Marcello
  Federico, Christian Federmann, Souhir Gahbiche, Hongyu Gong, Roman
  Grundkiewicz, Barry Haddow, Benjamin Hsu, D{\'a}vid Javorsk{\'y}, V{\u{e}}ra
  Kloudov{\'a}, Surafel Lakew, Xutai Ma, Prashant Mathur, Paul McNamee, Kenton
  Murray, Maria N{\v{a}}dejde, Satoshi Nakamura, Matteo Negri, Jan Niehues,
  Xing Niu, John Ortega, Juan Pino, Elizabeth Salesky, Jiatong Shi, Matthias
  Sperber, Sebastian St{\"u}ker, Katsuhito Sudoh, Marco Turchi, Yogesh Virkar,
  Alexander Waibel, Changhan Wang, and Shinji Watanabe. 2022.
\newblock \href {https://doi.org/10.18653/v1/2022.iwslt-1.10} {Findings of the
  {IWSLT} 2022 evaluation campaign}.
\newblock In \emph{Proceedings of the 19th International Conference on Spoken
  Language Translation (IWSLT 2022)}, pages 98--157, Dublin, Ireland (in-person
  and online). Association for Computational Linguistics.

\bibitem[{Arivazhagan et~al.(2020{\natexlab{a}})Arivazhagan, Cherry, I,
  Macherey, Baljekar, and Foster}]{ArivazhaganCTMB20}
Naveen Arivazhagan, Colin Cherry, Te~I, Wolfgang Macherey, Pallavi Baljekar,
  and George~F. Foster. 2020{\natexlab{a}}.
\newblock \href {https://doi.org/10.1109/ICASSP40776.2020.9054585}
  {Re-translation strategies for long form, simultaneous, spoken language
  translation}.
\newblock In \emph{2020 {IEEE} International Conference on Acoustics, Speech
  and Signal Processing, {ICASSP} 2020, Barcelona, Spain, May 4-8, 2020}, pages
  7919--7923. {IEEE}.

\bibitem[{Arivazhagan et~al.(2019)Arivazhagan, Cherry, Macherey, Chiu, Yavuz,
  Pang, Li, and Raffel}]{arivazhagan-etal-2019-monotonic}
Naveen Arivazhagan, Colin Cherry, Wolfgang Macherey, Chung-Cheng Chiu, Semih
  Yavuz, Ruoming Pang, Wei Li, and Colin Raffel. 2019.
\newblock \href {https://doi.org/10.18653/v1/P19-1126} {Monotonic infinite
  lookback attention for simultaneous machine translation}.
\newblock In \emph{Proceedings of the 57th Annual Meeting of the Association
  for Computational Linguistics}, pages 1313--1323, Florence, Italy.
  Association for Computational Linguistics.

\bibitem[{Arivazhagan et~al.(2020{\natexlab{b}})Arivazhagan, Cherry, Macherey,
  and Foster}]{arivazhagan-etal-2020-translation}
Naveen Arivazhagan, Colin Cherry, Wolfgang Macherey, and George Foster.
  2020{\natexlab{b}}.
\newblock \href {https://doi.org/10.18653/v1/2020.iwslt-1.27} {Re-translation
  versus streaming for simultaneous translation}.
\newblock In \emph{Proceedings of the 17th International Conference on Spoken
  Language Translation}, pages 220--227, Online. Association for Computational
  Linguistics.

\bibitem[{Bahar et~al.(2021)Bahar, Wilken, Di~Gangi, and
  Matusov}]{bahar-etal-2021-without}
Parnia Bahar, Patrick Wilken, Mattia~A. Di~Gangi, and Evgeny Matusov. 2021.
\newblock \href {https://doi.org/10.18653/v1/2021.iwslt-1.5} {Without further
  ado: Direct and simultaneous speech translation by {A}pp{T}ek in 2021}.
\newblock In \emph{Proceedings of the 18th International Conference on Spoken
  Language Translation (IWSLT 2021)}, pages 52--63, Bangkok, Thailand (online).
  Association for Computational Linguistics.

\bibitem[{Cettolo and Federico(2006)}]{Cettolo2006}
Mauro Cettolo and Marcello Federico. 2006.
\newblock \href {https://doi.org/10.1007/11816508\_66} {Text segmentation
  criteria for statistical machine translation}.
\newblock In \emph{In Proc. of Advances in Natural Language Processing,
  FinTAL}, volume 4139 of \emph{Lecture Notes in Computer Science}, pages
  664--673. Springer.

\bibitem[{Cho et~al.(2015)Cho, Niehues, Kilgour, and
  Waibel}]{cho-etal-2015-punctuation}
Eunah Cho, Jan Niehues, Kevin Kilgour, and Alex Waibel. 2015.
\newblock \href {https://aclanthology.org/2015.iwslt-papers.8} {Punctuation
  insertion for real-time spoken language translation}.
\newblock In \emph{Proceedings of the 12th International Workshop on Spoken
  Language Translation: Papers}, pages 173--179, Da Nang, Vietnam.

\bibitem[{Cho et~al.(2012)Cho, Niehues, and
  Waibel}]{cho-etal-2012-segmentationb}
Eunah Cho, Jan Niehues, and Alex Waibel. 2012.
\newblock \href {https://aclanthology.org/2012.iwslt-papers.15} {Segmentation
  and punctuation prediction in speech language translation using a monolingual
  translation system}.
\newblock In \emph{Proceedings of the 9th International Workshop on Spoken
  Language Translation: Papers}, pages 252--259, Hong Kong.

\bibitem[{Cho et~al.(2017)Cho, Niehues, and Waibel}]{Cho2017}
Eunah Cho, Jan Niehues, and Alex Waibel. 2017.
\newblock \href {https://doi.org/10.21437/Interspeech.2017-1320} {{NMT-Based
  Segmentation and Punctuation Insertion for Real-Time Spoken Language
  Translation}}.
\newblock In \emph{Proceedings of Interspeech 2017}, pages 2645--2649,
  Stockholm, Sweden. ISCA.

\bibitem[{Cho and Esipova(2016)}]{ChoE16}
Kyunghyun Cho and Masha Esipova. 2016.
\newblock \href {http://arxiv.org/abs/1606.02012v1} {Can neural machine
  translation do simultaneous translation?}
\newblock \emph{arXiv preprint arXiv:1606.02012v1}.

\bibitem[{Conneau et~al.(2020)Conneau, Khandelwal, Goyal, Chaudhary, Wenzek,
  Guzm{\'a}n, Grave, Ott, Zettlemoyer, and
  Stoyanov}]{conneau-etal-2020-unsupervised}
Alexis Conneau, Kartikay Khandelwal, Naman Goyal, Vishrav Chaudhary, Guillaume
  Wenzek, Francisco Guzm{\'a}n, Edouard Grave, Myle Ott, Luke Zettlemoyer, and
  Veselin Stoyanov. 2020.
\newblock \href {https://doi.org/10.18653/v1/2020.acl-main.747} {Unsupervised
  cross-lingual representation learning at scale}.
\newblock In \emph{Proceedings of the 58th Annual Meeting of the Association
  for Computational Linguistics}, pages 8440--8451, Online. Association for
  Computational Linguistics.

\bibitem[{Dai et~al.(2019)Dai, Yang, Yang, Carbonell, Le, and
  Salakhutdinov}]{dai-etal-2019-transformer}
Zihang Dai, Zhilin Yang, Yiming Yang, Jaime Carbonell, Quoc Le, and Ruslan
  Salakhutdinov. 2019.
\newblock \href {https://doi.org/10.18653/v1/P19-1285} {Transformer-{XL}:
  Attentive language models beyond a fixed-length context}.
\newblock In \emph{Proceedings of the 57th Annual Meeting of the Association
  for Computational Linguistics}, pages 2978--2988, Florence, Italy.
  Association for Computational Linguistics.

\bibitem[{Dyer et~al.(2013)Dyer, Chahuneau, and Smith}]{dyer-etal-2013-simple}
Chris Dyer, Victor Chahuneau, and Noah~A. Smith. 2013.
\newblock \href {https://aclanthology.org/N13-1073} {A simple, fast, and
  effective reparameterization of {IBM} model 2}.
\newblock In \emph{Proceedings of the 2013 Conference of the North {A}merican
  Chapter of the Association for Computational Linguistics: Human Language
  Technologies}, pages 644--648, Atlanta, Georgia. Association for
  Computational Linguistics.

\bibitem[{Elbayad et~al.(2020)Elbayad, Besacier, and Verbeek}]{elbayad20}
Maha Elbayad, Laurent Besacier, and Jakob Verbeek. 2020.
\newblock \href {https://doi.org/10.21437/Interspeech.2020-1241} {{Efficient
  Wait-k Models for Simultaneous Machine Translation}}.
\newblock In \emph{Proc. Interspeech 2020}, pages 1461--1465.

\bibitem[{Iranzo~Sanchez et~al.(2022)Iranzo~Sanchez, Civera, and
  Juan-C{\'\i}scar}]{iranzo-sanchez-etal-2022-simultaneous}
Javier Iranzo~Sanchez, Jorge Civera, and Alfons Juan-C{\'\i}scar. 2022.
\newblock \href {https://doi.org/10.18653/v1/2022.acl-long.480} {From
  simultaneous to streaming machine translation by leveraging streaming
  history}.
\newblock In \emph{Proceedings of the 60th Annual Meeting of the Association
  for Computational Linguistics (Volume 1: Long Papers)}, pages 6972--6985,
  Dublin, Ireland. Association for Computational Linguistics.

\bibitem[{Iranzo-S{\'a}nchez et~al.(2021)Iranzo-S{\'a}nchez, Civera~Saiz, and
  Juan}]{iranzo-sanchez-etal-2021-stream-level}
Javier Iranzo-S{\'a}nchez, Jorge Civera~Saiz, and Alfons Juan. 2021.
\newblock \href {https://doi.org/10.18653/v1/2021.findings-emnlp.58}
  {Stream-level latency evaluation for simultaneous machine translation}.
\newblock In \emph{Findings of the Association for Computational Linguistics:
  EMNLP 2021}, pages 664--670, Punta Cana, Dominican Republic. Association for
  Computational Linguistics.

\bibitem[{Iranzo-S{\'a}nchez et~al.(2020)Iranzo-S{\'a}nchez,
  Gim{\'e}nez~Pastor, Silvestre-Cerd{\`a}, Baquero-Arnal, Civera~Saiz, and
  Juan}]{iranzo-sanchez-etal-2020-direct}
Javier Iranzo-S{\'a}nchez, Adri{\`a} Gim{\'e}nez~Pastor, Joan~Albert
  Silvestre-Cerd{\`a}, Pau Baquero-Arnal, Jorge Civera~Saiz, and Alfons Juan.
  2020.
\newblock \href {https://doi.org/10.18653/v1/2020.emnlp-main.206} {Direct
  segmentation models for streaming speech translation}.
\newblock In \emph{Proceedings of the 2020 Conference on Empirical Methods in
  Natural Language Processing (EMNLP)}, pages 2599--2611, Online. Association
  for Computational Linguistics.

\bibitem[{Iranzo-Sánchez et~al.(2021)Iranzo-Sánchez, Jorge, Baquero-Arnal,
  Silvestre-Cerdà, Giménez, Civera, Sanchis, and Juan}]{IRANZOSANCHEZ2021}
Javier Iranzo-Sánchez, Javier Jorge, Pau Baquero-Arnal, Joan~Albert
  Silvestre-Cerdà, Adrià Giménez, Jorge Civera, Albert Sanchis, and Alfons
  Juan. 2021.
\newblock \href {https://doi.org/https://doi.org/10.1016/j.neunet.2021.05.013}
  {Streaming cascade-based speech translation leveraged by a direct
  segmentation model}.
\newblock \emph{Neural Networks}, 142:303--315.

\bibitem[{Kano et~al.(2021)Kano, Sudoh, and
  Nakamura}]{kano-etal-2021-simultaneous}
Yasumasa Kano, Katsuhito Sudoh, and Satoshi Nakamura. 2021.
\newblock \href {https://aclanthology.org/2021.wmt-1.120} {Simultaneous neural
  machine translation with constituent label prediction}.
\newblock In \emph{Proceedings of the Sixth Conference on Machine Translation},
  pages 1124--1134, Online. Association for Computational Linguistics.

\bibitem[{Koehn(2004)}]{koehn-2004-statistical}
Philipp Koehn. 2004.
\newblock \href {https://aclanthology.org/W04-3250} {Statistical significance
  tests for machine translation evaluation}.
\newblock In \emph{Proceedings of the 2004 Conference on Empirical Methods in
  Natural Language Processing}, pages 388--395, Barcelona, Spain. Association
  for Computational Linguistics.

\bibitem[{Koehn(2005)}]{koehn-2005-europarl}
Philipp Koehn. 2005.
\newblock \href {https://aclanthology.org/2005.mtsummit-papers.11} {{E}uroparl:
  A parallel corpus for statistical machine translation}.
\newblock In \emph{Proceedings of Machine Translation Summit X: Papers}, pages
  79--86, Phuket, Thailand.

\bibitem[{Kolss et~al.(2008)Kolss, Vogel, and Waibel}]{KolssVW08}
Muntsin Kolss, Stephan Vogel, and Alex Waibel. 2008.
\newblock \href
  {http://www.isca-speech.org/archive/interspeech\_2008/i08\_2735.html} {Stream
  decoding for simultaneous spoken language translation}.
\newblock In \emph{{INTERSPEECH} 2008, 9th Annual Conference of the
  International Speech Communication Association, Brisbane, Australia,
  September 22-26, 2008}, pages 2735--2738. {ISCA}.

\bibitem[{Kudo and Richardson(2018)}]{kudo-richardson-2018-sentencepiece}
Taku Kudo and John Richardson. 2018.
\newblock \href {https://doi.org/10.18653/v1/D18-2012} {{S}entence{P}iece: A
  simple and language independent subword tokenizer and detokenizer for neural
  text processing}.
\newblock In \emph{Proceedings of the 2018 Conference on Empirical Methods in
  Natural Language Processing: System Demonstrations}, pages 66--71, Brussels,
  Belgium. Association for Computational Linguistics.

\bibitem[{Li et~al.(2021{\natexlab{a}})Li, I, Arivazhagan, Cherry, and
  Padfield}]{Li2020b}
Daniel Li, Te~I, Naveen Arivazhagan, Colin Cherry, and Dirk Padfield.
  2021{\natexlab{a}}.
\newblock \href {https://doi.org/10.1109/ICASSP39728.2021.9413492} {Sentence
  boundary augmentation for neural machine translation robustness}.
\newblock In \emph{ICASSP 2021 - 2021 IEEE International Conference on
  Acoustics, Speech and Signal Processing (ICASSP)}, pages 7553--7557. {ISCA}.

\bibitem[{Li et~al.(2021{\natexlab{b}})Li, I, Arivazhagan, Cherry, and
  Padfield}]{li2021}
Daniel Li, Te~I, Naveen Arivazhagan, Colin Cherry, and Dirk Padfield.
  2021{\natexlab{b}}.
\newblock \href {https://doi.org/10.1109/ICASSP39728.2021.9413492} {Sentence
  boundary augmentation for neural machine translation robustness}.
\newblock In \emph{ICASSP 2021 - 2021 IEEE International Conference on
  Acoustics, Speech and Signal Processing (ICASSP)}, pages 7553--7557.

\bibitem[{Liu et~al.(2021)Liu, Du, Li, Li, and Chen}]{liu-etal-2021-cross}
Dan Liu, Mengge Du, Xiaoxi Li, Ya~Li, and Enhong Chen. 2021.
\newblock \href {https://doi.org/10.18653/v1/2021.emnlp-main.4} {Cross
  attention augmented transducer networks for simultaneous translation}.
\newblock In \emph{Proceedings of the 2021 Conference on Empirical Methods in
  Natural Language Processing}, pages 39--55, Online and Punta Cana, Dominican
  Republic. Association for Computational Linguistics.

\bibitem[{Ma et~al.(2019)Ma, Huang, Xiong, Zheng, Liu, Zheng, Zhang, He, Liu,
  Li, Wu, and Wang}]{ma-etal-2019-stacl}
Mingbo Ma, Liang Huang, Hao Xiong, Renjie Zheng, Kaibo Liu, Baigong Zheng,
  Chuanqiang Zhang, Zhongjun He, Hairong Liu, Xing Li, Hua Wu, and Haifeng
  Wang. 2019.
\newblock \href {https://doi.org/10.18653/v1/P19-1289} {{STACL}: Simultaneous
  translation with implicit anticipation and controllable latency using
  prefix-to-prefix framework}.
\newblock In \emph{Proceedings of the 57th Annual Meeting of the Association
  for Computational Linguistics}, pages 3025--3036, Florence, Italy.
  Association for Computational Linguistics.

\bibitem[{Ma et~al.(2020{\natexlab{a}})Ma, Zhang, and
  Zhou}]{ma-etal-2020-simple}
Shuming Ma, Dongdong Zhang, and Ming Zhou. 2020{\natexlab{a}}.
\newblock \href {https://doi.org/10.18653/v1/2020.acl-main.321} {A simple and
  effective unified encoder for document-level machine translation}.
\newblock In \emph{Proceedings of the 58th Annual Meeting of the Association
  for Computational Linguistics}, pages 3505--3511, Online. Association for
  Computational Linguistics.

\bibitem[{Ma et~al.(2020{\natexlab{b}})Ma, Pino, Cross, Puzon, and
  Gu}]{MaPCPG20}
Xutai Ma, Juan~Miguel Pino, James Cross, Liezl Puzon, and Jiatao Gu.
  2020{\natexlab{b}}.
\newblock Monotonic multihead attention.
\newblock In \emph{8th International Conference on Learning Representations,
  {ICLR} 2020, Addis Ababa, Ethiopia, April 26-30, 2020}.

\bibitem[{Matusov et~al.(2005)Matusov, Leusch, Bender, and
  Ney}]{matusov-etal-2005-evaluating}
Evgeny Matusov, Gregor Leusch, Oliver Bender, and Hermann Ney. 2005.
\newblock \href {https://aclanthology.org/2005.iwslt-1.19} {Evaluating machine
  translation output with automatic sentence segmentation}.
\newblock In \emph{Proceedings of the Second International Workshop on Spoken
  Language Translation}, Pittsburgh, Pennsylvania, USA.

\bibitem[{Papineni et~al.(2002)Papineni, Roukos, Ward, and
  Zhu}]{papineni-etal-2002-bleu}
Kishore Papineni, Salim Roukos, Todd Ward, and Wei-Jing Zhu. 2002.
\newblock \href {https://doi.org/10.3115/1073083.1073135} {{B}leu: a method for
  automatic evaluation of machine translation}.
\newblock In \emph{Proceedings of the 40th Annual Meeting of the Association
  for Computational Linguistics}, pages 311--318, Philadelphia, Pennsylvania,
  USA. Association for Computational Linguistics.

\bibitem[{Post(2018)}]{post-2018-call}
Matt Post. 2018.
\newblock \href {https://doi.org/10.18653/v1/W18-6319} {A call for clarity in
  reporting {BLEU} scores}.
\newblock In \emph{Proceedings of the Third Conference on Machine Translation:
  Research Papers}, pages 186--191, Brussels, Belgium. Association for
  Computational Linguistics.

\bibitem[{Pu et~al.(2021)Pu, Chung, Parikh, Gehrmann, and
  Sellam}]{pu-etal-2021-learning}
Amy Pu, Hyung~Won Chung, Ankur Parikh, Sebastian Gehrmann, and Thibault Sellam.
  2021.
\newblock \href {https://doi.org/10.18653/v1/2021.emnlp-main.58} {Learning
  compact metrics for {MT}}.
\newblock In \emph{Proceedings of the 2021 Conference on Empirical Methods in
  Natural Language Processing}, pages 751--762, Online and Punta Cana,
  Dominican Republic. Association for Computational Linguistics.

\bibitem[{Scherrer et~al.(2019)Scherrer, Tiedemann, and
  Lo{\'a}iciga}]{scherrer-etal-2019-analysing}
Yves Scherrer, J{\"o}rg Tiedemann, and Sharid Lo{\'a}iciga. 2019.
\newblock \href {https://doi.org/10.18653/v1/D19-6506} {Analysing concatenation
  approaches to document-level {NMT} in two different domains}.
\newblock In \emph{Proceedings of the Fourth Workshop on Discourse in Machine
  Translation (DiscoMT 2019)}, pages 51--61, Hong Kong, China. Association for
  Computational Linguistics.

\bibitem[{Schneider and Waibel(2020)}]{schneider-waibel-2020-towards}
Felix Schneider and Alexander Waibel. 2020.
\newblock \href {https://doi.org/10.18653/v1/2020.iwslt-1.28} {Towards stream
  translation: Adaptive computation time for simultaneous machine translation}.
\newblock In \emph{Proceedings of the 17th International Conference on Spoken
  Language Translation}, pages 228--236, Online. Association for Computational
  Linguistics.

\bibitem[{Sen et~al.(2022)Sen, Bojar, and Haddow}]{sukanta2022}
Sukanta Sen, Ondřej Bojar, and Barry Haddow. 2022.
\newblock \href {https://arxiv.org/abs/2210.09754v1} {Simultaneous translation
  for unsegmented input: A sliding window approach}.
\newblock \emph{arXiv preprint arXiv:2210.09754v1}.

\bibitem[{Stolcke and Shriberg(1996)}]{Stolcke1996}
Andreas Stolcke and Elizabeth Shriberg. 1996.
\newblock \href {http://www.isca-speech.org/archive/icslp\_1996/i96\_1005.html}
  {Automatic linguistic segmentation of conversational speech}.
\newblock In \emph{Proceedings of the Fourth International Conference on Spoken
  Language Processing}, volume~2, pages 1005--1008. {ISCA}.

\bibitem[{Tiedemann(2009)}]{Tiedemann09}
J{\"o}rg Tiedemann. 2009.
\newblock \emph{News from OPUS - A Collection of Multilingual Parallel Corpora
  with Tools and Interfaces}, volume~V.

\bibitem[{Tiedemann and Scherrer(2017)}]{tiedemann-scherrer-2017-neural}
J{\"o}rg Tiedemann and Yves Scherrer. 2017.
\newblock \href {https://doi.org/10.18653/v1/W17-4811} {Neural machine
  translation with extended context}.
\newblock In \emph{Proceedings of the Third Workshop on Discourse in Machine
  Translation}, pages 82--92, Copenhagen, Denmark. Association for
  Computational Linguistics.

\bibitem[{Vaswani et~al.(2017)Vaswani, Shazeer, Parmar, Uszkoreit, Jones,
  Gomez, Kaiser, and Polosukhin}]{Vaswani2017}
Ashish Vaswani, Noam Shazeer, Niki Parmar, Jakob Uszkoreit, Llion Jones,
  Aidan~N Gomez, \L~ukasz Kaiser, and Illia Polosukhin. 2017.
\newblock \href
  {https://proceedings.neurips.cc/paper_files/paper/2017/file/3f5ee243547dee91fbd053c1c4a845aa-Paper.pdf}
  {Attention is all you need}.
\newblock In \emph{Advances in Neural Information Processing Systems},
  volume~30. Curran Associates, Inc.

\bibitem[{Wilken et~al.(2020)Wilken, Alkhouli, Matusov, and
  Golik}]{wilken-etal-2020-neural}
Patrick Wilken, Tamer Alkhouli, Evgeny Matusov, and Pavel Golik. 2020.
\newblock \href {https://doi.org/10.18653/v1/2020.iwslt-1.29} {Neural
  simultaneous speech translation using alignment-based chunking}.
\newblock In \emph{Proceedings of the 17th International Conference on Spoken
  Language Translation}, pages 237--246, Online. Association for Computational
  Linguistics.

\bibitem[{Yao and Haddow(2020)}]{yao-haddow-2020-dynamic}
Yuekun Yao and Barry Haddow. 2020.
\newblock \href {https://aclanthology.org/2020.amta-research.12} {Dynamic
  masking for improved stability in online spoken language translation}.
\newblock In \emph{Proceedings of the 14th Conference of the Association for
  Machine Translation in the Americas (Volume 1: Research Track)}, pages
  123--136, Virtual. Association for Machine Translation in the Americas.

\bibitem[{Zhang et~al.(2021)Zhang, Titov, Haddow, and
  Sennrich}]{zhang-etal-2021-beyond}
Biao Zhang, Ivan Titov, Barry Haddow, and Rico Sennrich. 2021.
\newblock \href {https://doi.org/10.18653/v1/2021.acl-long.200} {Beyond
  sentence-level end-to-end speech translation: Context helps}.
\newblock In \emph{Proceedings of the 59th Annual Meeting of the Association
  for Computational Linguistics and the 11th International Joint Conference on
  Natural Language Processing (Volume 1: Long Papers)}, pages 2566--2578,
  Online. Association for Computational Linguistics.

\bibitem[{Zhang et~al.(2022)Zhang, He, Wu, and Wang}]{zhang-etal-2022-learning}
Ruiqing Zhang, Zhongjun He, Hua Wu, and Haifeng Wang. 2022.
\newblock \href {https://doi.org/10.18653/v1/2022.acl-long.542} {Learning
  adaptive segmentation policy for end-to-end simultaneous translation}.
\newblock In \emph{Proceedings of the 60th Annual Meeting of the Association
  for Computational Linguistics (Volume 1: Long Papers)}, pages 7862--7874,
  Dublin, Ireland. Association for Computational Linguistics.

\bibitem[{Zhang et~al.(2020)Zhang, Zhang, He, Wu, and
  Wang}]{zhang-etal-2020-learning-adaptive}
Ruiqing Zhang, Chuanqiang Zhang, Zhongjun He, Hua Wu, and Haifeng Wang. 2020.
\newblock \href {https://doi.org/10.18653/v1/2020.emnlp-main.178} {Learning
  adaptive segmentation policy for simultaneous translation}.
\newblock In \emph{Proceedings of the 2020 Conference on Empirical Methods in
  Natural Language Processing (EMNLP)}, pages 2280--2289, Online. Association
  for Computational Linguistics.

\bibitem[{Zheng et~al.(2020{\natexlab{a}})Zheng, Liu, Zheng, Ma, Liu, and
  Huang}]{zheng-etal-2020-simultaneous}
Baigong Zheng, Kaibo Liu, Renjie Zheng, Mingbo Ma, Hairong Liu, and Liang
  Huang. 2020{\natexlab{a}}.
\newblock \href {https://doi.org/10.18653/v1/2020.acl-main.254} {Simultaneous
  translation policies: From fixed to adaptive}.
\newblock In \emph{Proceedings of the 58th Annual Meeting of the Association
  for Computational Linguistics}, pages 2847--2853, Online. Association for
  Computational Linguistics.

\bibitem[{Zheng et~al.(2019)Zheng, Ma, Zheng, and
  Huang}]{zheng-etal-2019-speculative}
Renjie Zheng, Mingbo Ma, Baigong Zheng, and Liang Huang. 2019.
\newblock \href {https://doi.org/10.18653/v1/D19-1144} {Speculative beam search
  for simultaneous translation}.
\newblock In \emph{Proceedings of the 2019 Conference on Empirical Methods in
  Natural Language Processing and the 9th International Joint Conference on
  Natural Language Processing (EMNLP-IJCNLP)}, pages 1395--1402, Hong Kong,
  China. Association for Computational Linguistics.

\bibitem[{Zheng et~al.(2020{\natexlab{b}})Zheng, Yue, Huang, Chen, and
  Birch}]{Zheng2020b}
Zaixiang Zheng, Xiang Yue, Shujian Huang, Jiajun Chen, and Alexandra Birch.
  2020{\natexlab{b}}.
\newblock \href {https://doi.org/10.24963/ijcai.2020/551} {Towards making the
  most of context in neural machine translation}.
\newblock In \emph{Proceedings of the Twenty-Ninth International Joint
  Conference on Artificial Intelligence}, pages 3983--3989, Yokohama, Yokohama,
  Japan. International Joint Conferences on Artificial Intelligence
  Organization.

\end{thebibliography}
\bibliographystyle{acl_natbib}

%
%
% APPENDIX AFTER THIS
%
%

\appendix

\onecolumn

\section{Appendix: Additional Figures}

\begin{figure*}[htb]
        \centering
        \includegraphics[width=.355\textwidth]{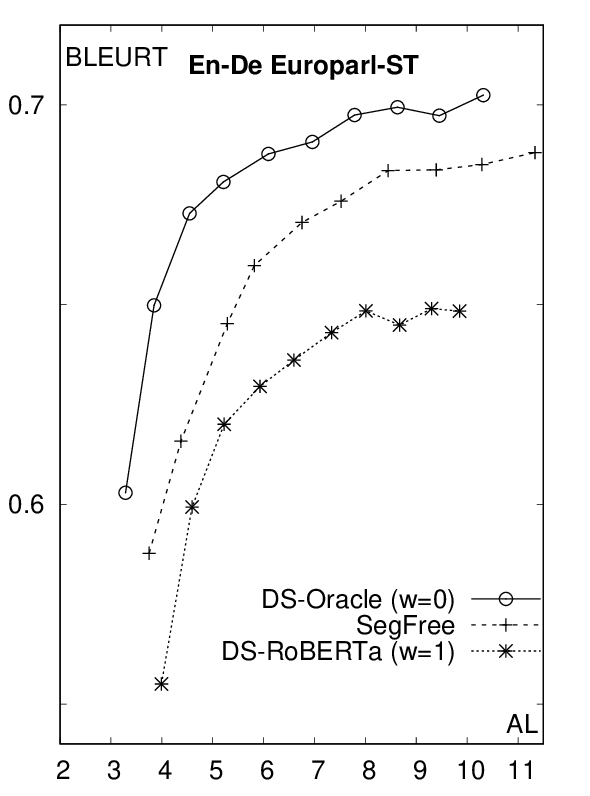}\hspace*{-5mm}
        \includegraphics[width=.355\textwidth]{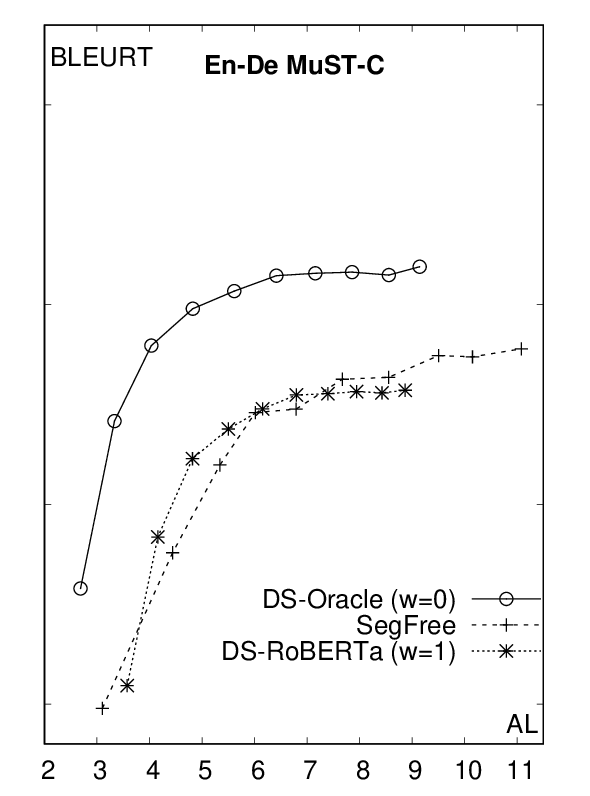}\hspace*{-5mm}
        \includegraphics[width=.355\textwidth]{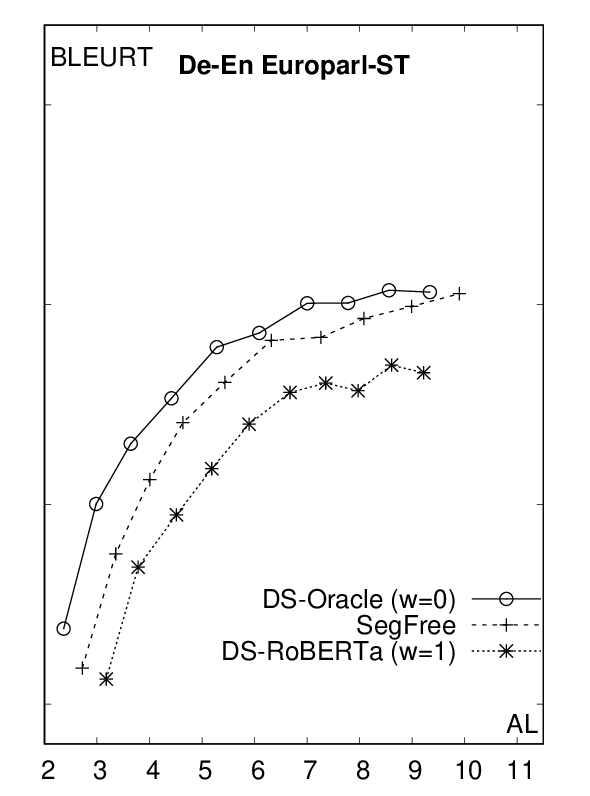}
        \caption{BLEURT vs. AL on the English to German Europarl-ST
          (left) and MuST-C (center) test sets, and the German to
          English Europarl-ST (right) test set. WindowRetrans curves
          are not shown for the sake of clarity, as they are
          significantly lower than the
          rest.\label{fig:test_combined_BLEURT}}
\end{figure*}

\begin{figure*}[htb]
        \centering
        \includegraphics[width=.455\textwidth]{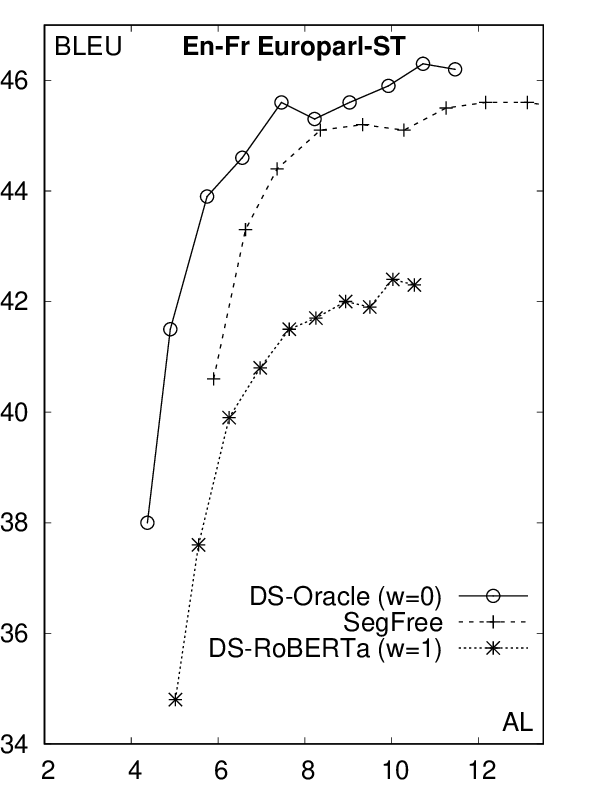}\hspace*{-5mm}
        \includegraphics[width=.455\textwidth]{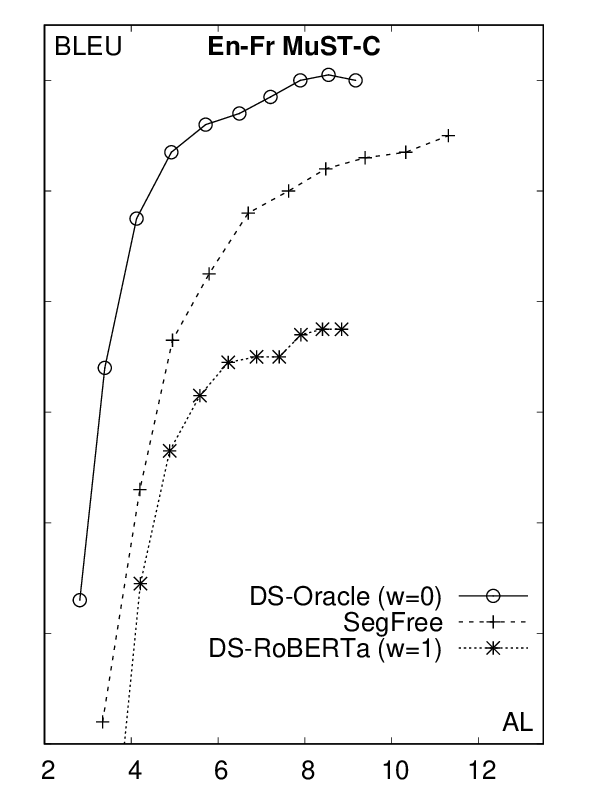}\hspace*{-5mm}
        \caption{BLEU vs. AL on the English to French Europarl-ST
	  (left) and MuST-C (right) test sets. WindowRetrans curves
	  are not shown for the sake of clarity, as they are
	  significantly lower than the
	  rest. \label{fig:test_combined_enfr}}
\end{figure*}

\begin{figure*}[htb]
        \centering
        \includegraphics[width=.455\textwidth]{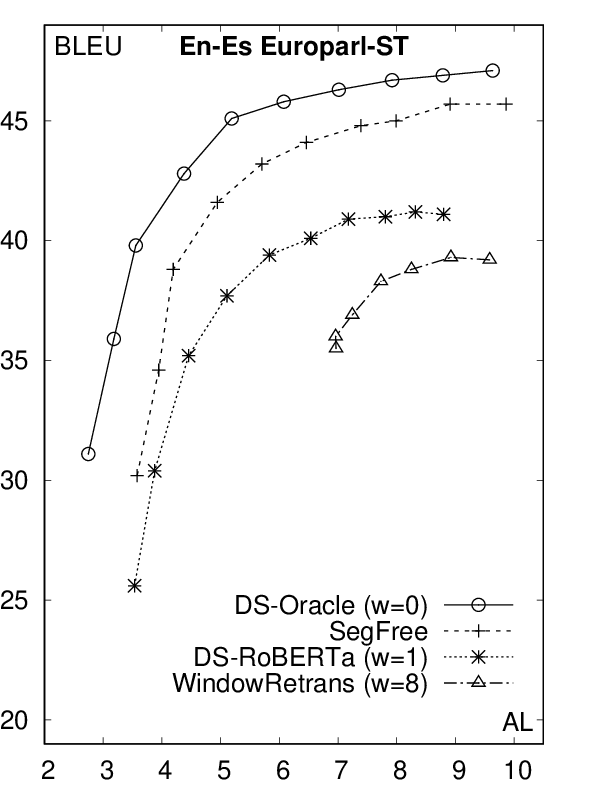}\hspace*{-5mm}
        \includegraphics[width=.455\textwidth]{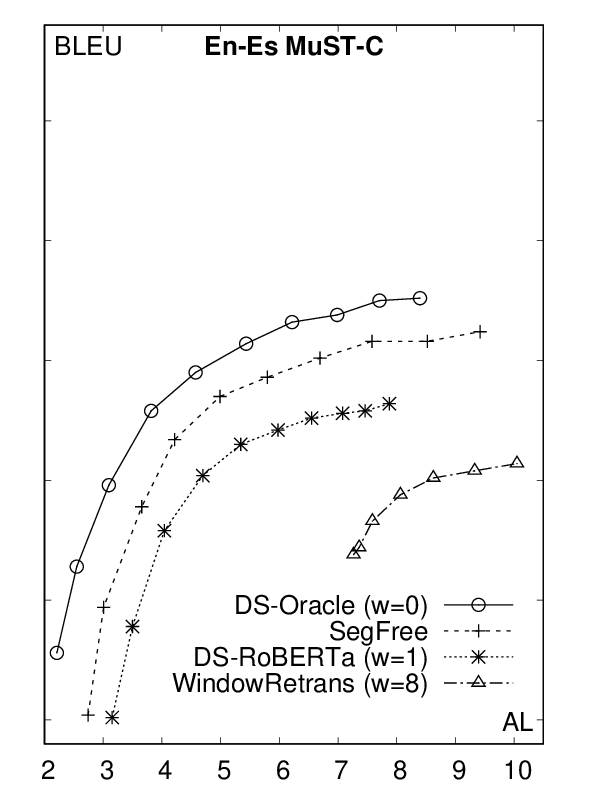}\hspace*{-5mm}        
        \caption{BLEU vs. AL on the English to Spanish Europarl-ST
          (left) and MuST-C (right) test sets.
          set.\label{fig:test_combined_enes}}
\end{figure*}

\begin{figure*}[htb]
        \centering
        \includegraphics[width=.455\textwidth]{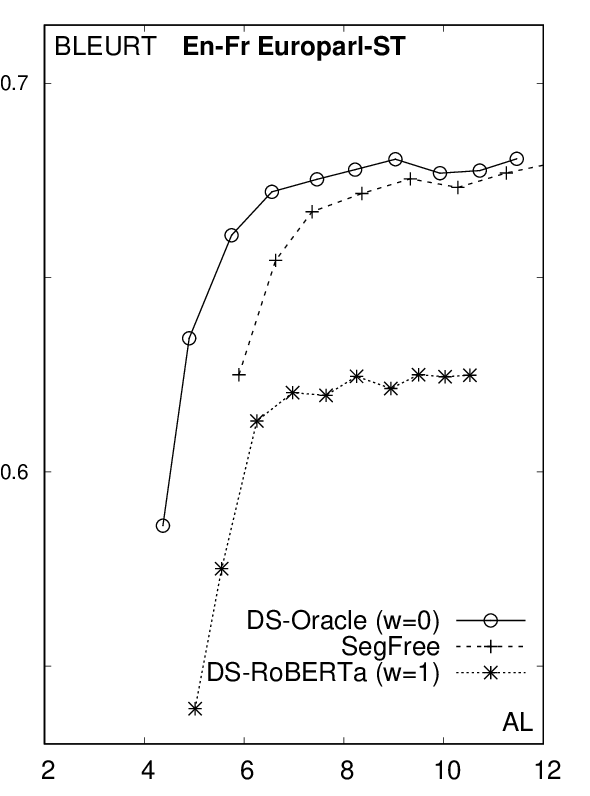}\hspace*{-5mm}
        \includegraphics[width=.455\textwidth]{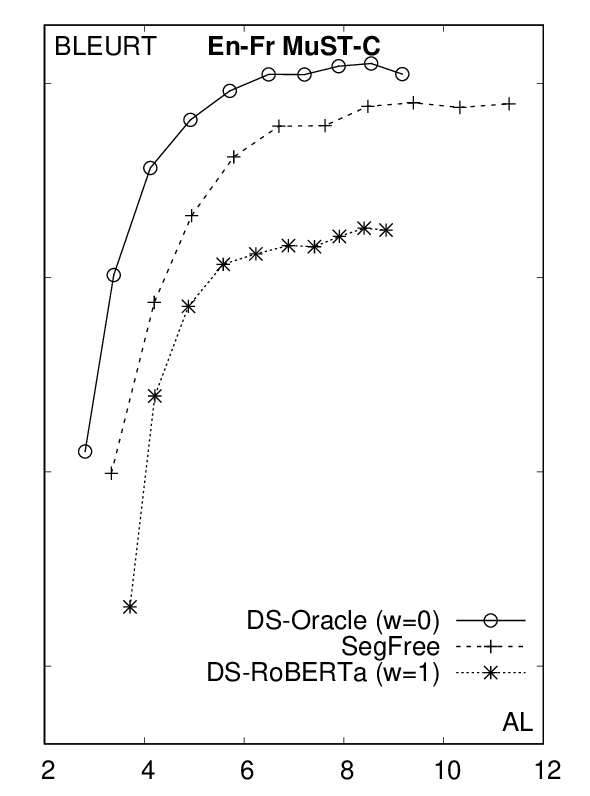}\hspace*{-5mm}        
        \caption{BLEURT vs. AL on the English to French Europarl-ST
	  (left) and MuST-C (right) test sets. WindowRetrans curves
	  are not shown for the sake of clarity, as they are
	  significantly lower than the
	  rest. \label{fig:test_combined_enfr_BLEURT}}
\end{figure*}

\begin{figure*}[htb]
        \centering
        \includegraphics[width=.455\textwidth]{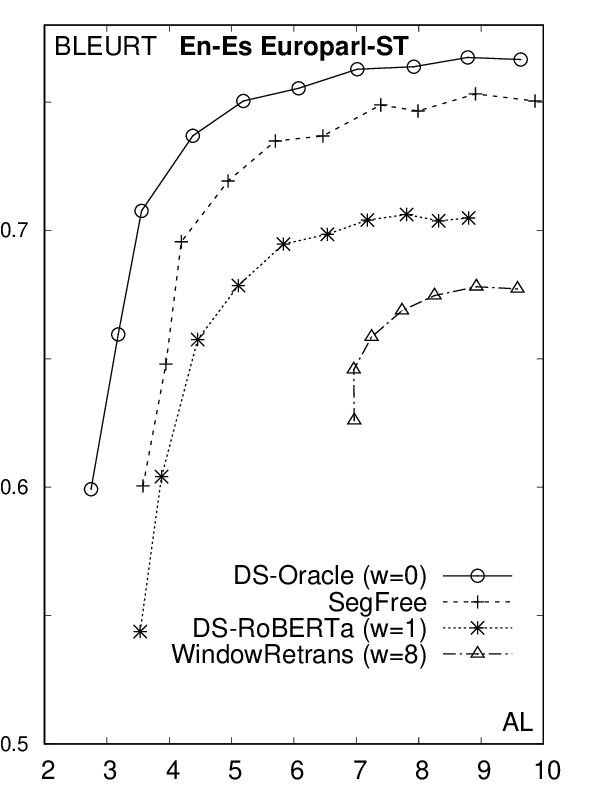}\hspace*{-5mm}
        \includegraphics[width=.455\textwidth]{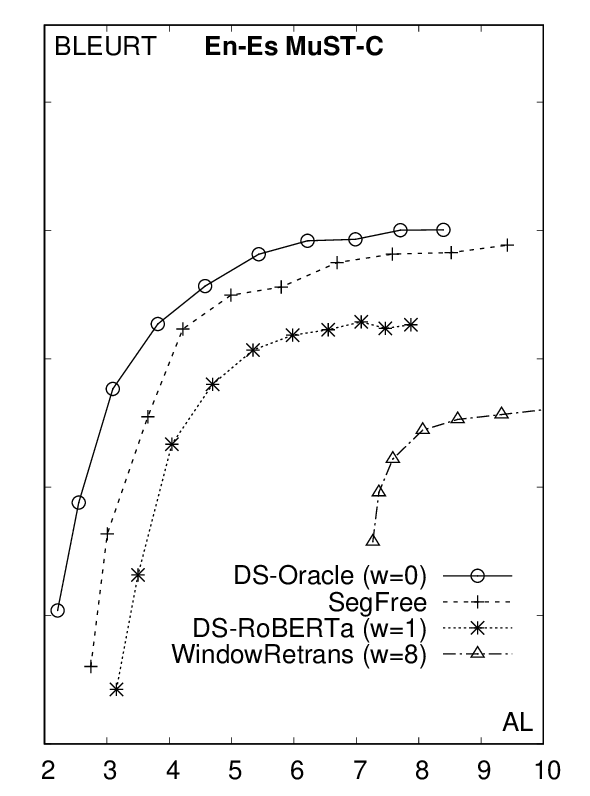}\hspace*{-5mm}        
        \caption{BLEURT vs. AL on the English to Spanish Europarl-ST
          (left) and MuST-C (right) test
          sets. \label{fig:test_combined_enes_BLEURT}}
\end{figure*}

%\begin{figure*}[htb]
%        \centering
%        \includegraphics[width=.455\textwidth]{fig/Europarl-ST_test_enes_tl.eps}\hspace*{-5mm}
%        \includegraphics[width=.455\textwidth]{fig/MuST-C_test_enes_tl.eps}\hspace*{-5mm}
%        \caption{BLEU vs. Translation Lag on the English to Spanish Europarl-ST
%          (left) and MuST-C (right) test sets.
%          set.\label{fig:test_combined_enes_tl}}
%\end{figure*}

%\begin{figure*}[htb]
%        \centering
%        \includegraphics[width=.455\textwidth]{fig/Europarl-ST_test_enfr_tl.eps}\hspace*{-5mm}
%        \includegraphics[width=.455\textwidth]{fig/MuST-C_test_enfr_tl.eps}\hspace*{-5mm}
%        \caption{BLEU vs. Translation Lag on the English to French Europarl-ST
%          (left) and MuST-C (right) test sets.
%          set.\label{fig:test_combined_enes_tl}}
%\end{figure*}

\end{document}